\documentclass[10pt, twoside, twocolumn]{IEEEtran}

\usepackage{times}
\usepackage{epsfig}
\usepackage{graphicx}
\usepackage{amsmath}
\usepackage{amssymb}
\usepackage{helvet}
\usepackage{courier}
\usepackage{graphicx}
\usepackage{subfigure}
\usepackage{color}
\usepackage{amssymb}
\usepackage{algorithm}
\usepackage{algorithmic}
\usepackage{multirow}
\usepackage{xcolor}
\usepackage{multirow}
\usepackage{longtable}

\def\para#1{\medskip\noindent{\bf #1}}

\begin{document}

\title{Bayesian Neighbourhood Component Analysis}

\author{Dong~Wang and Xiaoyang~Tan \thanks{Dong Wang and Xiaoyang Tan are with the
    Department of Computer Science and Technology, Nanjing University
    of Aeronautics and Astronautics, P.R.~China. Corresponding author: Xiaoyang Tan
    (x.tan@nuaa.edu.cn).}}


\maketitle

\begin{abstract}
Learning a good distance metric in feature space potentially improves the performance of the KNN classifier and is useful in many real-world applications. Many metric learning algorithms are however based on the point estimation of a quadratic optimization problem, which is time-consuming, susceptible to overfitting, and lack a natural mechanism to reason with parameter uncertainty, an important property useful especially when the training set is small and/or noisy. To deal with these issues, we present a novel Bayesian metric learning method, called Bayesian NCA, based on the well-known Neighbourhood Component Analysis method, in which the metric posterior is characterized by the local label consistency constraints of observations, encoded with a similarity graph instead of independent pairwise constraints. For efficient Bayesian optimization, we explore the variational lower bound over the log-likelihood of the original NCA objective. Experiments on several publicly available datasets demonstrate that the proposed method is able to learn robust metric measures from small size dataset and/or from challenging training set with labels contaminated by errors. The proposed method is also shown to outperform a previous pairwise constrained Bayesian metric learning method.
\end{abstract}

\section{Introduction}
Learning a good distance metric in feature space is crucial in many real-world applications. It has been shown to significantly improve the performance of object classification \cite{mensink2012metric}, image retrieval \cite{hoi2006learning}, image ranking \cite{li2015ordinal}, face identification \cite{guillaumin2010multiple}, kinship verification \cite{lu2014neighborhood}, clustering \cite{ye2007adaptive}, or person re-identification \cite{dikmen2011pedestrian}.
Most of distance metric learning (DML) methods aim to learn a linear transformation which pulls together samples from the same class while pushing away those from different classes.

There has been considerable research on distance metric learning over the past few years \cite{guillaumin2009you} \cite{guillaumin2009tagprop} \cite{koestinger2012large} \cite{bian2012constrained} \cite{fouad2013incorporating} \cite{shen2014efficient} \cite{li2015distributed}.
Although metric learning algorithms have achieved great success, there are several challenges remaining to be addressed: Firstly, most algorithms are based on point estimation, which is sensitive to the choice of training examples; Secondly, they tend to be over-fitting especially
when training set is small, partly due to the large parameter space; Last but not least, learning a Mahalanobis metric usually involves very big computational complexity ( i.e., $o(N^3)$, $N$ is the number of training data). The recently proposed pairwise constrained Bayesian metric learning method (BML) \cite{Yang07} overcomes some of the above issues by taking the prior distribution of the transformation matrix into account. However, it treats each sample independently and ignores the different importance of each sample, which limits its efficiency in learning.

Graph constraints can be regarded as the extension to pairwise constraints (\cite{Yang07}, \cite{koestinger2012large}) by simultaneously building the similarity constraints between one sample with many others. One typical metric learning method using graph constraints is the well-known Neighbourhood Component Analysis (NCA) \cite{goldberger2004neighbourhood}, which is conceptually simple and is developed under a well-formulated probabilistic framework. There are several extensions of this method in literatures, such as  the maximum-margin nearest neighbor (LMNN) \cite{weinberger2005distance},  Nearest Class Mean \cite{mensink2012metric}, label noise robust NCA \cite{wang2014robust}, and so on. However, all of them are based on the point estimation.

Another problem with many existing metric learning methods is that they seldom take the influence of label noise into consideration. The errors in data labels may result in unnecessary complex model which in turn leads to overfitting. This problem has been previously studied under the umbrella of agnostic learning \cite{kearns1994toward}, in which the relationship between the label and the data is largely relaxed. Since then many label noise robust methods have been proposed \cite{lawrence2001estimating} \cite{cantador2005boosting} \cite{van2010novel} \cite{leung2011handling} \cite{fefilatyev2012label} \cite{frenay2013classification}, but they usually have high computation cost and do not focus on metric learning.

In this paper we present a graph constrained Bayesian metric learning method to address all the above issues. The method is based on the NCA method but for the first time, extends it under the Bayesian framework, hence called Bayesian Neighbourhood Component Analysis (BNCA). Unlike previous studies on metric learning methods, our method naturally takes account the uncertainty due to the unreliable parameter estimation under difficult conditions and is less susceptible to overfitting by exploiting the prior knowledge. Compared to other methods, it also provides more robust estimation even when there are errors in data labels, with the help of effective group label consistency constraints. Furthermore, we develop a variational lower bound of the log-likelihood of the objective, which significantly reduces the computational cost while preserving the effectiveness of Bayesian estimation. Finally, we give a throughout analysis of the proposed method from the angle of sample selection, robustness property, and computational complexity.

To verify the effectiveness of the proposed method, we have conducted a comprehensive study that compares the BNCA with other DML techniques on several real-world applications, including image classification, digital recognition, and face recognition. Our experiments demonstrate that the BNCA method is able to learn robust metric measures from small size dataset and/or from unperfect training set with its data labels contaminated by errors. It is also shown to outperform a previous pairwise constrained Bayesian metric learning method and several other state of the art DML methods.

The remaining parts of this paper are organized as follows: In Section \ref{sec_prelimiaries}, preliminaries are provided regarding the closely related  metric learning algorithms, then we detail our proposed Beyesian Neighbourhood Component Analysis in Section \ref{sec_proposedmethod}. In section \ref{sec_anl} we give a thorough analysis of this method. In section \ref{sec_exp}, we investigate the performance of our method empirically over several popular datasets. We conclude this paper in Section \ref{sec_conclude}.

\section{Preliminary}\label{sec_prelimiaries}
Assuming that we have a dataset $D$ of $N$ data points, denoted as $D=\{x_i,~y_i\},i=1,2,3,...,N$, where $y_i$ is the label of the $i$-th data point $x_i$. In distance metric learning, we aim to learn a Mahalanobis matrix---$A$ using some form of supervision information. Mahalanobis distance metric measures the squared distance between
two data points $x_i$ and $x_j$ as follows,
\begin{equation}\label{eq_dA2}
d_{A}^2(x_i,x_j) = (x_i-x_j)^TA(x_i-x_j)
\end{equation}
where $A\ge0$ is a positive semidefinite matrix and $x_i,x_j\in R^d$ is a pair of samples $(i, j)$. For simplicity we denote $d_{A}^2(x_i,x_j)$ as $d_{A^{ij}}^2$. With these notations, in what follows, we give a brief overview on two state-of-the-art works closely related to ours in learning a Mahalanobis metric, i.e., NCA \cite{goldberger2004neighbourhood} and pairwise constrained Bayesian metric learning \cite{Yang07}.

\subsection{Neighbourhood Component Analysis}\label{sec_NCA}
The NCA algorithm \cite{goldberger2004neighbourhood} begins by constructing a complete graph with each data point as its node. Let the weight of each edge between any two nodes denoted as $p_{ij}$. It is interpreted as the probability that data point $x_{i}$ selects $x_{j}$ as its neighbor and can be calculated as follows,
\begin{equation}\label{fuc_=pij}
p_{ij} = \frac{\exp(-d_{A^{ij}}^{2})}{\sum_{t\in N_i} \exp(-d_{A^{it}}^2)}
\end{equation}
where $N_i$ denotes the set of neighbors of $x_i$. It can be checked that $p_{ij}\geq 0$ and $\sum_{j\in N_i} p_{ij}=1$, and hence $p_{ij}$ is a valid probability measure.

The object of NCA is then to learn a linear transformation $A$ which maximizes the log likelihood that after transformation each data point selects the points with the same labels as itself as neighbors, i.e.,
\begin{equation}\label{fuc_=logA}
\max~L(A)= \sum_{i}\log(\sum_{j\in N_i} 1\{y_i=y_j\}\cdot p_{ij}) \\
\end{equation}

\subsection{Pairwise Constrained Bayesian Metric Learning}
Yang et al. \cite{Yang07} proposed a Bayesian metric learning (BML) method that estimates the posterior distribution for the distance metric from labeled pairwise constraints. It defines the probability for two data points
$x_i$ and $x_j$ to form an equivalence or inequivalence constraint under a given distance metric A:
\begin{eqnarray}\label{eq_bml}
  P(y_{ij}|x_i,x_j,A,\mu) &=& \frac{1}{1+\exp(y_{ij}(d_{A^{ij}}^2-\mu))} \\
  where \qquad y_{ij}&=&
\begin{cases}
+1& \text{$(x_i,x_j)\in \mathcal{S}$}\\
-1& \text{$(x_i,x_j)\in \mathcal{D}$}
\end{cases}
\end{eqnarray}

In the above, $\mathcal{S}$ and $\mathcal{D}$ respectively denote the sets of equivalence or inequivalence constraints. Given this, the posterior distribution of metric $A$ and the threshold $\mu$ can be estimated by maximizing the following objective:
\begin{equation}\label{eq_BMLlikelihood}
L(A,\mu)= \prod_{(i,j)}P(y_{ij}|x_i,x_j,A,\mu)p(A)p(\mu)
\end{equation}

This method effectively overcomes some of the limitations of traditional metric learning methods. However, it does not take the structure of data into consideration and does not scale well. Particularly, since its objective just requires that the distance between similar pairs of points should be lower than that between dissimilar ones, all pairs $(i,j)$ ($o(N^2)$) need to be calculated for training. This not only increases the computational cost, but also ignores the importance weight of each sample regards to model training, which significantly decreases the learning efficiency because ideally we should focus more on those data whose labels are not consistent with most of its neighbors, instead of treating them indifferently. This problem is partially addressed later by Yang et al. with an active learning method for data pair selection \cite{Yang07}, but the computational cost remains high.

\section{Bayesian Neighbourhood Component Analysis}\label{sec_proposedmethod}
\subsection{The Proposed Method}
We start our derivation by considering the three components of a general Bayesian model, i.e., prior, likelihood, posterior. Since the original NCA is a discriminant model, we write its likelihood as $P(Y|X,A)$ in our Bayesian NCA, where $A$ is the linear transformation matrix to be learnt. We follow the same assumption as that of NCA, i.e., the sample labels are conditionally independent given the labels of their nearest neighbors, hence the conditional model can be written as
\begin{equation}\label{eq_likelihood}
P(Y|X,A) = \frac{1}{Z(A)}\prod_iP(y_i|x_i,Y_{N_i},X_{N_i},A)
\end{equation}
where $Z(A)$ is a normalizing constant known as the partition function. To be consistent with NCA, we define
\begin{equation}\label{eq_likli1}
P(y_i=k|x_i,Y_{N_i},X_{N_i},A) = \frac{\sum_{j\in N_i}1\{y_j=k\}\cdot \exp(-d_{A^{ij}}^{2})}{\sum_{t\in N_i} \exp(-d_{A^{it}}^2)}\\
\end{equation}
Comparing eq.~(\ref{eq_likli1}) with eq.~(\ref{eq_bml}), we see that one of the major differences between our model and the BML lies in that the local Neighbourhood structure $N_i$ is naturally embedded into the model in our method.

To compute the posterior of the distance metric $A$, a prior for it should be specified, and a convenient choice for this could be the Wishart prior. Unfortunately, it is well-known that combining the Gaussian prior with a non-Gaussian likelihood is difficult to compute. In addition, the integration of $A$ is untractable as well.

To bypass the above issues, we first approximate the distance metric $A$ as a linear combination of the top eigenvectors of the observed data, and then estimate the posterior distribution of the combination weights using variational method.

\subsubsection{Eigen Approximation}
Let $X$ = $(x_1,x_2,...,x_N)$ denote all the examples, and $v_l,(l=1,2,...,d)$ be the top $d$ eigenvectors of $XX^T$. Inspired by \cite{Yang07}, we approximate $A$ using the first $d$ eigenvectors, i.e., $A$= $\sum_{l=1}^d\gamma_lv_lv_l^T$, where $\gamma_l,(l=1,2,...,d)$ are the combination coefficients. With this,  the likelihood $P(y_i=k|x_i,Y_{N_i},X_{N_i},A)$ in eq.~(\ref{eq_likli1}) reduces to its equivalent form $P(y_i=k|x_i,Y_{N_i},X_{N_i},\gamma)$:
\begin{equation}\label{eq_relikli1}
\begin{split}
P(y_i=k|x_i,Y_{N_i},X_{N_i},\gamma) = \frac{\sum_{j\in N_i}1\{y_j=k\}\cdot \exp(-d_{\gamma^{ij}}^{2})}{\sum_{t\in N_i} \exp(-d_{\gamma^{it}}^{2}))}
\end{split}
\end{equation}
where we define:
\begin{equation}\label{eq_defw}
\begin{split}
w_{ij}^l &= (v_l^T(x_i-x_j))^2\\
w_{ij} &= [w_{ij}^1, w_{ij}^2, ..., w_{ij}^d]^T\\
\gamma &=[\gamma_1, \gamma_2,..., \gamma_d]^T
\end{split}
\end{equation}
then $d_{\gamma^{ij}}^{2} = d_{A^{ij}}^{2} = \gamma^Tw_{ij}$.

Our task then boils down to compute the posterior distribution of $\gamma$. For simplicity, we assume that the prior distribution of $\gamma$ to be Gaussian:
\begin{equation}\label{eq_propri}
  p(\gamma) = N(\gamma|m_0,V_0)
\end{equation}
where $m_0$ and $V_0$ are respectively mean and covariance.
\subsubsection{Variational Approximation}\label{sec_VA}
At the second step we employ the variational method to estimate the posterior distribution of $\gamma$. The main idea is to introduce variational distributions for $\gamma$ to construct the lower bound and then maximize the
the lower bound to obtain the approximate estimation for the posterior distribution. We begin with the unnormalized logarithm likelihood $\log \{Z(A)P(Y|X,\gamma)\}$ . Note that maximizing this objective directly regarding to $A$ leads to the standard NCA algorithm, but our goal here is for local variational approximation, hence the partition function $Z(A)$ is simply treated as a constant.  Particularly,
\begin{equation}\label{eq_vi}
\begin{split}
L&= \log \{Z(A)P(Y|X,\gamma)\}\\
=& \sum_i \sum_k1\{y_i =k\}\log \{p(y_i=k|x_i,Y_{N_i},X_{N_i},\gamma)\}\\
=& \sum_i \sum_k 1\{y_i =k\}\log\{ \frac{\sum_{j\in N_i}1\{y_j=k\}\cdot \exp(-d_{\gamma^{ij}}^{2})}{\sum_{t\in N_i} \exp(-d_{\gamma^{it}}^2)}\}
\end{split}
\end{equation}
Since $\log(a+b)>\log(a)+\log(b)$ if $0<a,b<1$, we have,
\begin{equation}\label{eq_vi}
\begin{split}
&L > \sum_i \sum_{j\in N_i}y_{ij} \log \{\frac{\exp(-d_{\gamma^{ij}}^{2})}{\sum_{t\in N_i} \exp(-d_{\gamma^{it}}^2)}\}\\
&> \sum_i \sum_{j\in N_i}y_{ij} \log\{\frac{1}{1+\sum_{t\in N_i} \exp(d_{\gamma^{ij}}^2-d_{\gamma^{it}}^2)}\}\\
\end{split}
\end{equation}
Let $x_{N_{i_1}},x_{N_{i_2}},...,x_{N_{i_K}}$ be respectively the $K$ nearest neighbors of $x_i$. For convenience, we introduce the following notations:
\begin{equation}\label{eq_defWeta}
\begin{split}
  &\eta_{ij}^t = d_{\gamma^{ij}}^2-d_{\gamma^{it}}^2 =(w_{ij}-w_{it})^T\gamma\\
  &W_i^j = [w_{ij}-w_{iN_{i_1}},w_{ij}-w_{iN_{i_2}},...,w_{ij}-w_{iN_{i_K}}]\\
  &\eta_{ij}=[\eta_{ij}^{N_{i_1}},\eta_{ij}^{N_{i_2}},...,\eta_{ij}^{N_{i_K}}]^T =(W_i^j)^T\gamma
\end{split}
\end{equation}
Recall the definition of log-sum-exp function:
\begin{equation}
\begin{split}
lse(\eta_{ij}) \triangleq \log(1+\sum_{t\in N_i} \exp(\eta_{ij}^t))\\
\end{split}
\end{equation}
Then eq.~(\ref{eq_vi}) can be rewritten as:
\begin{equation}
\begin{split}
L > -\sum_i \sum_{j\in N_i} y_{ij} lse(\eta_{ij})
\end{split}
\end{equation}
Using Bohning's quadratic bound (see \cite{murphy2012machine} page 758, section.~21.8.2 for details) we have:
\begin{equation}\label{eq_Bohning}
\begin{split}
&L > \sum_i \sum_{j\in N_i} y_{ij} \{-\frac{1}{2}\eta_{ij}^TH\eta_{ij}+b_{ij}^T \eta_{ij} -c_{ij}\}\\
&=\sum_i \sum_{j\in N_i} y_{ij}\{-\frac{1}{2}\gamma^T W_i^jH(W_i^j)^T\gamma+b_{ij}^T(W_i^j)^T\gamma-c_{ij}\}
\end{split}
\end{equation}
where  $c_{ij}$ is a constant and the remaining notations are defined as,
\begin{equation}\label{eq_defH}
\begin{split}
&H = \frac{1}{2}[I_K-\frac{1}{K+1}1_K 1_K^T]
\end{split}
\end{equation}
\begin{equation}\label{eq_defb}
\begin{split}
&b_{ij} = H\psi_{ij}-g(\psi_{ij})\\
\end{split}
\end{equation}
\begin{equation}
\begin{split}
&g(\psi_{ij}) = \exp(\psi_{ij}-lse(\psi_{ij}))
\end{split}
\end{equation}
Note that $\psi_{ij}$ is the variational parameter.

Now we proceed to compute the posterior distribution of $\gamma$, which we model as a Gaussian, denoted as $N(\gamma|m_T,V_T)$. We write the unconstrained posterior distribution,
\begin{equation}
\begin{split}
p(\gamma|X,Y) \propto p(Y|X,\gamma)p(\gamma),
\end{split}
\end{equation}
and plug in the approximated likelihood (\ref{eq_Bohning}) and the prior distribution $N(\gamma|m_0,V_0)$ to get,
\begin{equation}\label{eq_Vt}
\begin{split}
&V_T = [V_0^{-1}+\sum_i\sum_{j\in N_i}y_{ij} W_i^jH(W_i^j)^T]^{-1}\\
\end{split}
\end{equation}
\begin{equation}\label{eq_mt}
\begin{split}
&m_T = V_T(V_0^{-1}m_0+\sum_i\sum_{j\in N_i}y_{ij} W_i^jb_{ij})\\
\end{split}
\end{equation}
Finally  the variational parameter $\psi_{ij}$ is updated as,
\begin{equation}\label{eq_psi}
\begin{split}
\psi_{ij} = (W_i^j)^Tm_T
\end{split}
\end{equation}

We summarize the proposed method in Algorithm.~\ref{alg_bnca}.
\begin{algorithm}[htb] 
\renewcommand{\algorithmicrequire}{\textbf{Input:}}
\renewcommand\algorithmicensure {\textbf{Output:} }
\caption{Bayesian Neighbourhood Component Analysis.}
\label{alg:Framwork}
\begin{algorithmic}[1] 
\REQUIRE ~~\\ 
Input: Training set $\{(x_{i},y_{i})|~i=1,2,...,N\}$, prior distribution $N(\gamma|m_0,V_0)$ ;\\
\ENSURE ~~\\ 
posterior distribution $N(\gamma|m_T,V_T)$\\
------ Training Stage
\STATE Define $W_i^j$, $H$ according to (\ref{eq_defWeta}) and (\ref{eq_defH}) respectively.
\STATE Compute $V_T$ with eq.~(\ref{eq_Vt}).
\STATE Repeat
\STATE ~~~~compute $\psi_{ij}$ for all $(i,j)$ with eq.~(\ref{eq_psi})\\
\STATE ~~~~compute $b_{ij}$ for all $(i,j)$ with eq.~(\ref{eq_defb})\\
\STATE ~~~~compute $m_T$ with Eq.~(\ref{eq_mt}). \\
\STATE Until converged.
\STATE Return $N(\gamma|m_T,V_T)$.
\end{algorithmic} \label{alg_bnca}
\end{algorithm}

\subsection{Distance Estimation}\label{sec_infbnca}
For inference we are interested in the expectation of the point-to-point distance $d_{\gamma^{ij}}^2$ for a new couple of data $(i,j)$ according to the posterior distribution of $\gamma$, which is a Gaussian distribution as shown above. Particularly, we have,
\begin{equation}\label{eq_dij p}
\begin{split}
d_{\gamma^{ij}}^2 \sim N(d_{\gamma^{ij}}^2|m_{ij}, \sigma_{ij}^2)
\end{split}
\end{equation}
where
\begin{equation}\label{eq_pdij}
\begin{split}
m_{ij} &= (w_{ij})^Tm_T\\
\sigma_{ij}^2 & = (w_{ij})^TV_Tw_{ij}
\end{split}
\end{equation}
It is worthwhile to mention that this mechanism of outputting model uncertainty in distance metric calculation is potentially beneficial to many real world applications but unfortunately is largely ignored in the field. For example, in the application of image retrieval rather than ranking the results purely based on the estimated similarity, we could now construct a more robust ranking scheme by taking the value of the related similarity uncertainty (i.e., $\sigma_{ij}^2$) into account. We would not pursue this issue any further as it is out of the range of this paper, but it will be the focus of our future work.

Instead, one could simply use the MAP value---$m_{ij}$ to estimate each $d_{\gamma^{ij}}^2$. To see the difference between this with the traditional NCA method, we decompose its expectation as follows,
\begin{equation}\label{eq_dijd}
\begin{split}
E(d_{\gamma^{ij}}^2) &= \sum_l w_{ij}^lm_T^l\\
&= \sum_l(x_i-x_j)^Tv_lm_T^lv_l^T(x_i-x_j)
\end{split}
\end{equation}
where $w_{ij}^l$ and $m_T^l$ are the $l$-th element of $w_{ij}$ and $m_T$ respectively.

Now defining the new coordinate axes as $[v'_1,v'_2,...,v'_d]$ with $v'_l = (m_T^l)^{\frac{1}{2}}\cdot v_l$, we see that the inference equation~(\ref{eq_pdij}) essentially calculates the distance in a feature space spanned by the top $d$ the eigenvectors of $XX^T$ but scaled  by $(m_T)^{\frac{1}{2}}$, according to the distribution of the corresponding eigenvalues (c.f., Fig.~\ref{fig:bnca}).

\subsection{Prediction under Parameter Uncertainty}\label{sec_uncertainty}
Under the difficult condition of small size training samples or samples with label noise, a single estimate of parameter $A$ tends to be unreliable and the traditional DML methods that are based on it may cause overconfidence in the future predictions. In other words, they just make predictions but cannot tell whether these predictions make sense. By contrast, for Bayesian methods this is really not a problem because no errors would be introduced due to the inaccurate estimation of $A$. Particularly, the prediction for a never-seen sample $x_i$ can be obtained from $p(y_i|x_i, Y_{N_i},X)$. Recall that the variational posterior of metric parameter $\gamma$ is a Gaussian distribution, i.e., $q(\gamma)= N(\gamma|m^T,V^T)$ (c.f., eq.~(\ref{eq_mt}) $\&$ eq.~(\ref{eq_Vt})), we have,
\begin{equation}\label{eq_uncertainty}
\begin{split}
p(y_i|x_i, Y_{N_i},X_{N_i}) &= \int_{\gamma}p(y_i|x_i, Y_{N_i},X_{N_i},\gamma)q(\gamma)d\gamma  \\
\end{split}
\end{equation}
The difficulty here is that this integration of $\gamma$ is untractable, because $p(y_i|x_i,Y_{N_i},X,\gamma)$ is a multinomial distribution while $q(\gamma)$ is a Gaussian one. Instead we adopt a MCMC method \cite{andrieu2003introduction} to approximate this expectation:
\begin{equation}\label{eq_uncertainty_mcmc}
\begin{split}
p(y_i|x_i,Y_{N_i},X_{N_i}) &\approx \frac{1}{T}\sum_{l=1}^T p(y_i|x_i,Y_{N_i},X_{N_i},\gamma_l) \\
\end{split}
\end{equation}
where $\gamma_l(l=1,2,...,T)$ are sampled i.i.d. from $q(\gamma)$.

\section{Analysis of the Proposed Method}\label{sec_anl}
\subsection{Adaptive Sample Selection in Learning}\label{sec_secln}
 In the process of metric learning, it is beneficial to exploit the local property of samples in the input space to improve the learning efficiency. Take the NCA algorithm as an example. Its gradient is calculated as follows,
\begin{equation}\label{eq_ncagrad}
\begin{split}
&\frac{\partial{L}}{\partial{A}} = 2A\sum_i(\sum_{j\in N_i}p_{ij}x_{ij}x_{ij}^T-\frac{\sum_{j\in N_i}y_{ij} p_{ij}x_{ij}x_{ij}^T}{\sum_{j\in N_i}y_{ij} p_{ij}})
\end{split}
\end{equation}
That is, for any point $x_i$, if and only if all its $K$ nearest neighbors have the same labels as that of $x_i$, then $\sum_{j\in N_i}y_{ij} p_{ij}=1$, which means that the gradient equals to $\vec{0}$. Hence the NCA algorithm would pay more attention on those points whose labels are inconsistent with its $K$ nearest neighbors. In other words, not all pairs $(x_i,x_j)$ are active in constraining the search space of the transformation matrix in the same way.

Similar observations can be made in other NCA extensions, such as the Large Margin NN (LMNN \cite{weinberger2005distance}) method:
\begin{equation}\label{eq_lmnn}
\min~L(A) = \sum_i\sum_{j\in N_i}(d_{\gamma^{ij}}^2 + \mu\sum_{l \in N_i}(1-y_{il})\xi_{ijl}(A))
\end{equation}
where the first term can be seen as a regularization while the second one penalizes those data points that violate the large margin condition.

But the situation is different for pairwise constrained models, e.g., \cite{Yang07}, in the sense that they usually lack an automatic sample selection mechanism in the metric learning objective by itself (c.f., eq.~(\ref{eq_BMLlikelihood})). In our opinion, it is important to give different importance weights to different points during training, because doing this properly potentially allows us to significantly reduce the computational costs and to lessen the likelihood of overfitting. Actually to avoid computing distance for all possible data pairs, Yang et al. \cite{Yang07} designed an effective active learning method to select the most uncertainty pairs in training process, but the algorithm still needs to compute and store all the pairs' uncertainty scores.
\begin{figure}[t!]
\centering
\includegraphics[width=0.95\linewidth]{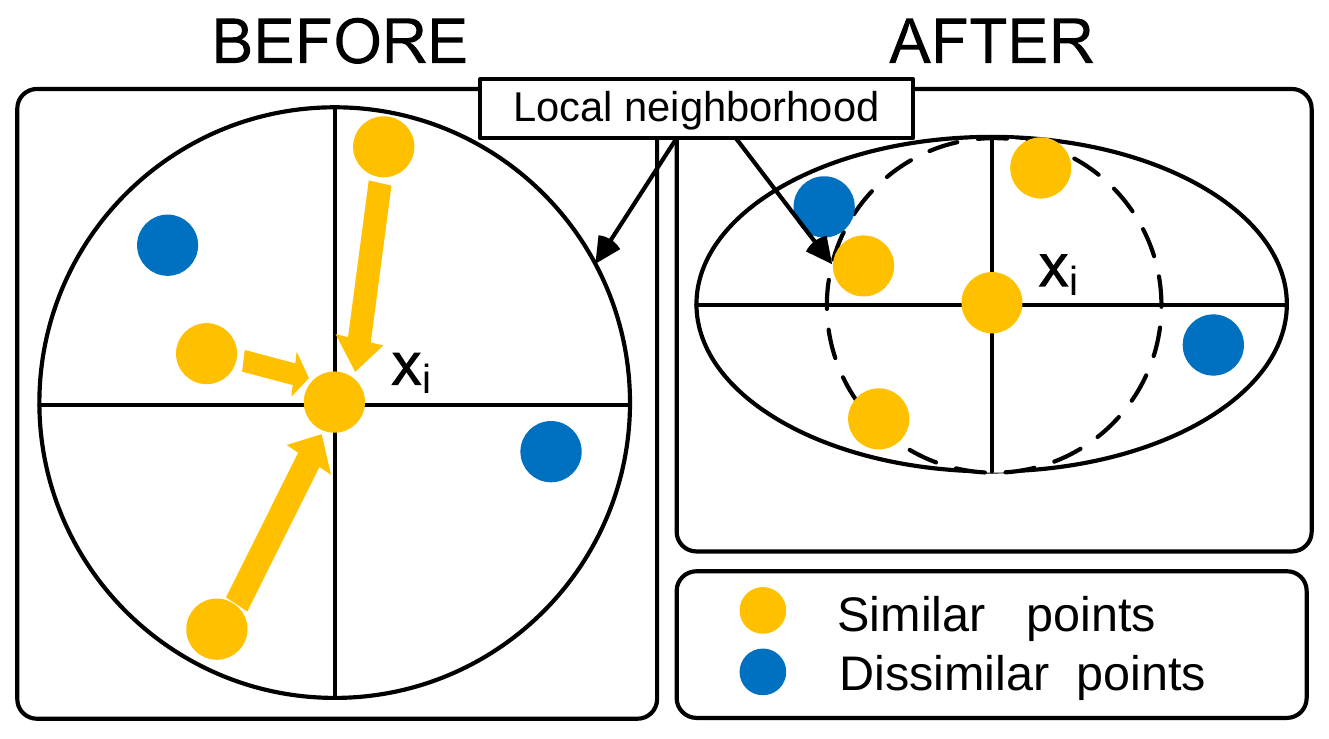}
\caption{BNCA appropriately scale the axis to allow $x_i$'s neighbors in same class to be closer.}
\label{fig:bnca}
\end{figure}

Let's come to our results of BNCA shown in eq.~(\ref{eq_Vt}) and eq.~(\ref{eq_mt}). For simplicity here we only care about the diagonal elements $V_T^{ll}~(l = 1, 2, ..., d)$ of $V_T$. Let us define $W_i^{j^{l}}$ as the $l$-th row of $W_i^{j}$ ($W_i^{j} = [W_i^{j^{1}},W_i^{j^{2}},...,W_i^{j^{d}}]^T$):
\begin{equation}
\begin{split}
W_i^{j^{l}} = [w_{ij}^l-w_{iN_{i1}}^l,w_{ij}^l-w_{iN_{i2}}^l,...,w_{ij}^l-w_{iN_{iK}}^l]
\end{split}
\end{equation}
From eq.~(\ref{eq_Vt}) we get:
\begin{equation}
\begin{split}
&V_T^{ll} = ((V_0^{ll})^{-1}+\sum_i\sum_{j\in N_i}\sum_{t\in N_i}y_{ij}W_i^{j^{l}}H_{ll}(W_i^{j^{l}})^T)^{-1}
\end{split}
\end{equation}
Assume that $K\gg1$, $H$ can be approximated by $\frac{1}{2}I$, such that:
\begin{equation}
\begin{split}
&V_T^{ll} = ((V_0^{ll})^{-1}+\frac{1}{2}\sum_i\sum_{j\in N_i}y_{ij}\sum_{t\in N_i}(w_{ij}^l-w_{it}^l)^2)^{-1}\label{eq_vii}
\end{split}
\end{equation}

If we simply throw away the non-diag elements of $V_T$ from Eq.~(\ref{eq_mt}), we see that $V_T^{ll}$ is in proportion to $m_T^l$. In other words, in BNCA we scale the axis $v_l$ by reducing the variance of $\gamma_l$ such that all $x_i$'s neighbors in the same class will be closer in that direction, as illustrated in Fig.~\ref{fig:bnca}.

To see how our proposed BNCA handles different data points adaptively, we consider the following two extreme circumstances: 1) all $x_i$'s nearest neighbors have the same labels as that of $x_i$ and have the same distance to $x_i$; 2) none of $x_i$'s nearest neighbors belongs to the same class of $x_i$. In both cases the term $\sum_{j\in N_i}y_{ij}\sum_{t\in N_i}(w_{ij}^l-w_{it}^l)^2$ (c.f., eq.~(\ref{eq_vii})) equals to 0, such that the variance won't be changed by $x_i$. In the first circumstance those $x_i$ can be thought of as perfect points that needn't to be adjusted, while the second case illustrates how our method handles the data in a robust way when some of them lie on the decision boundary or when their labels are too noisy to be learnt from.

\subsection{Robustness against Label Noise}\label{sec_any_ln}
To reveal the influence of label noise on the training of a DML model, we start the analysis with the NCA method. First let us denote the two major components of its gradient (eq.~(\ref{eq_ncagrad})) as $C_E$ and $C_I$, respectively,
\begin{align}\label{fuc_CI}
C_E=&\sum_{i}\sum_{j\in N_i} p_{ij}x_{ij}x_{ij}^T\\
C_I=& \sum_i\frac{\sum_{j\in N_i}y_{ij} p_{ij}x_{ij}x_{ij}^T}{\sum_{j\in N_i}y_{ij} p_{ij}}\\
=& \sum_{i}\sum_j p_{ij}x_{ij}x_{ij}^T\sum_k\frac{1(y_i=k)\cdot 1(y_j=k)}{\sum_j1(y_j=k)\cdot p_{ij}}
\end{align}

We see that,
\begin{equation}
\frac{\partial L}{\partial A}=2A(C_E-C_I)
\label{eq_gradient2}
\end{equation}

Intuitively, the $C_E$ term denotes the total scatter  matrix of the data points lying on the manifold induced by $A$ and $C_I$ is the corresponding intra-class scatter matrix eq.~(\ref{eq_gradient2}) reveals that, up to a constant matrix, in each step the NCA algorithm tries to seek a better linear transformation such that after projection the total covariance becomes 'larger' while the intra-class covariance becomes 'smaller'. However, when the class labels are inaccurate or noisy, the estimation of $C_I$ tends to be inaccurate (the $C_E$ will be not influenced by this).

The same situation occurs in LMNN. As can be seen from eq.~(\ref{eq_lmnn}), label noise would possibly result in a lot of incorrect training triples $(ijl)$,  which pull together samples from different class while pushing away those from the same classes. This issue becomes more and more troublesome with the increase in the noise level, and actually all the DML techniques trained in a supervised way would suffer from this if not properly taken care of. This is witnessed by our experiments given later, showing that under some high noise level, many traditional start-of-the-art DML methods such as NCA and LMNN will even be inferior to the unsupervised baseline, i.e., PCA.

Finally let us come back to the proposed BNCA. From eq.~(\ref{eq_Vt}) and eq.~(\ref{eq_mt}) we see that the estimated posterior distribution is influenced by the label noise as well, which may lead to inaccurate scaling of the projection axis in each training iteration (c.f., Algorithm.~\ref{alg_bnca}). But in our BNCA method there are two built-in mechanisms that help to alleviate this problem: the first one is due to the introducing of prior knowledge, which works like a regularization to the model explanation of the data. The other one can be seen from the variational inference result of eq.~(\ref{eq_vii}), where the influence of data points with label noise is effectively reduced because these erroneous data tend to have quite different labels with their neighbors. It is worth mentioning that in BNCA there is no any pre-processing of label correction and hence no extra time cost would be involved for this.

\subsection{Computational Cost}\label{sec_anltime}
To analyze the computational cost of the proposed method, first note that usually the most time-consuming step in Laplace approximation or a conjugate gradient method is related to the calculation of Hessian matrix. In each iteration, it needs $o(N^3)$ computational cost (where $N$ is the number of training data, e.g., if $N$ is $10^{3}$, the computational cost could be as large as $10^{9}$). While in our case, thanks to the Bohning's approximation the Hessian matrix becomes a constant matrix (c.f., eq.~(\ref{eq_defH})), calculated only once.

Furthermore, in NCA and other point based metric learning methods their objective are commonly optimized using some gradient descent based method, such as stochastic gradient descent \cite{bottou2010large} or conjugate gradient descend \cite{shewchuk1994introduction}, etc.. Those methods usually need a number of iterations to perform local search. Even worse, when encountering non-convex problems, they usually need more time to converge and may get lost in the local minimums. By contrast our BNCA method avoids the time-consuming gradient iterations completely - the lower bound of eq.~(\ref{eq_Bohning})) actually gives us an analytic solution (c.f., eq.~(\ref{eq_Vt}) and eq.~(\ref{eq_mt})).
As a consequence, our computational cost is significantly reduced compared to others.

\section{Experiments}\label{sec_exp}

To verify the effectiveness of the proposed method, in this section we first compare the robustness performance of our method with several related DML methods on the datasets either with small sample size or with label noise, then we turn to investigate in depth the behavior of the proposed method.

\subsection{Experimental Settings}\label{sec_imple}
We compare the performance of the proposed method with several other closely related DML methods, including NCA \cite{goldberger2004neighbourhood}, LMNN \cite{weinberger2005distance}, metric learning for Nearest Class Mean \cite{mensink2012metric} (NCM), and pairwise constrained Bayesian metric learning \cite{Yang07} (BML). Both the NCA and the LMNN are metric learning methods with graph constraints, and the NCA is the method our method based on, hence chosen to be the baseline algorithm. Like ours, the BML method proposed by Yang et al. \cite{Yang07} is a Bayesian method as well, but with pairwise constraints for learning. Finally, we also adopted an unsupervised latent feature learning method, i.e., Principal Component Analysis \cite{wold1987principal} (PCA) for comparison, since it is completely irrelevant to the issue of label noise.

For all the methods except the NCM (which has its own classifier), we used the K-NN method equipped with the corresponding learnt metric for classification. In addition, the performance of all the compared methods are based on the original implementation kindly provided by the corresponding authors, and the related hyper-parameters are fine-tuned through cross validation. Each experiment is repeated for ten times, and both the mean and the standard deviation of the classification accuracy are reported. To evaluate the performance of the compared methods, we also conducted pairwise one-tail statistical test under significance level 0.05.

\para{Implementation Details}
In Bayesian NCA there are a few parameters need to be initialized, mainly including the parameters of the prior distribution $N(\gamma|m_0, V_0)$ and local variational approximation parameters , including $b_{ij}$ (eq.~(\ref{eq_defb})) and $\psi_{ij}$ (eq.~(\ref{eq_psi})). In this work we set $m_0$ to $\epsilon$$\vec{1}$ where $\vec{1}$ is all 1's vector and $\epsilon$ is a small scalar (e.g. 0.1). From section \ref{sec_infbnca} we see that this choice of $m_0$ is equivalent to initialize BNCA with PCA, which is commonly used in metric learning for initialization and will not be affected by label noise. Besides, we set $V_0$ to $\sigma I$, where $\sigma$ is a vary small value (e.g. $0.001$). This helps to preserve the stability of $V_T$ (eq.~(\ref{eq_Vt})), one important property related to overfitting. Then we compute $b_{ij}$ and $\psi_{ij}$ according to eq.~(\ref{eq_defb}) and eq.~(\ref{eq_psi}) respectively.

\begin{table*}[t!]
\caption{Comparative performance (\%) on UCI datasets with varying sizes of training set. (The asterisks indicate a statistically significant difference between the
second best performer and the proposed method at a significance level of 0.05.)}
\begin{center}
\begin{tabular}{c|c|l|l|l|l|l|l}
\hline
Dataset & training Size$(\#)$ & PCA & NCA & NCM  &LMNN &BML  &BNCA\\
\hline
\hline
\multirow{3}{*}{Balance} & 10 &69.42$\pm$ 0.20 &67.85$\pm$ 0.35 &66.28$\pm$ 0.20 &70.38$\pm$ 0.33 &71.38$\pm$ 0.33$^{\ast}$ &\textbf{75.42$\pm$ 0.31}\\
                        & 20 &74.53$\pm$ 0.12 &74.38$\pm$ 0.29 &72.62$\pm$ 0.13 &77.38$\pm$ 0.28$^{\ast}$ &76.74$\pm$ 0.26 &\textbf{80.76$\pm$ 0.26}\\
                        & 30 &77.23$\pm$ 0.10 &79.38$\pm$ 0.18 &76.16$\pm$ 0.10 &81.18$\pm$ 0.16$^{\ast}$ &79.74$\pm$ 0.14 &\textbf{83.26$\pm$ 0.14}\\
\hline
\hline
\multirow{3}{*}{Ionosphere} &10 &68.91$\pm$ 0.17 &70.97$\pm$ 0.28 &68.91$\pm$ 0.17 &73.54$\pm$ 0.30 &73.68$\pm$ 0.28$^{\ast}$ &\textbf{76.75$\pm$ 0.28}\\
                            &20 &73.12$\pm$ 0.15 &75.75$\pm$ 0.21 &73.19$\pm$ 0.15 &78.64$\pm$ 0.23$^{\ast}$ &76.52$\pm$ 0.20 &\textbf{81.22$\pm$ 0.21}\\
                            &30 &76.14$\pm$ 0.11 &80.96$\pm$ 0.17 &77.82$\pm$ 0.10 &83.88$\pm$ 0.16$^{\ast}$ &80.86$\pm$ 0.16 &\textbf{85.96$\pm$ 0.16}\\
\hline
\hline
\multirow{3}{*}{Spambase} &10 &73.42$\pm$ 0.23 &70.38$\pm$ 0.37 &70.08$\pm$ 0.24  &72.85$\pm$ 0.35 &75.38$\pm$ 0.35$^{\ast}$ &\textbf{79.42$\pm$ 0.34}\\
                         &20 &77.53$\pm$ 0.20  &76.74$\pm$ 0.24 &75.56$\pm$ 0.20 &79.38$\pm$ 0.25$^{\ast}$ &78.52$\pm$ 0.24 &\textbf{83.76$\pm$ 0.24}\\
                         &30 &79.53$\pm$ 0.14  &81.74$\pm$ 0.23 &79.74$\pm$ 0.14 &84.38$\pm$ 0.23$^{\ast}$ &80.97$\pm$ 0.22 &\textbf{86.76$\pm$ 0.22}\\
\hline
\end{tabular}
\end{center}
\label{tb:uci}
\end{table*}

\begin{table*}[t!]
\caption{Comparative performance (\%) on different datasets with varying degree of label noise.(The asterisks indicate a statistically significant difference between the
second best performer and the proposed method at a significance level of 0.05.)}
\begin{center}
\begin{tabular}{c|c|l|l|l|l|l|l}
\hline
Dataset & noise level$(\%)$ & PCA & NCA & NCM  &LMNN &BML  &BNCA\\
\hline
\hline
\multirow{4}{*}{Caltech-10} &0 &80.12$\pm$ 0.12  &81.49$\pm$ 0.20  &76.65$\pm$ 0.20 &82.84$\pm$ 0.21$^{\ast}$ &80.68$\pm$ 0.20 &\textbf{83.89$\pm$ 0.20}\\
                           &10 &79.02$\pm$ 0.14  &79.03$\pm$ 0.22  &74.49$\pm$ 0.20 &79.38$\pm$ 0.23$^{\ast}$ &79.11$\pm$ 0.22 &\textbf{81.79$\pm$ 0.22}\\
                           &20 &76.15$\pm$ 0.14  &74.59$\pm$ 0.26  &68.97$\pm$ 0.20 &74.27$\pm$ 0.28 &76.53$\pm$ 0.26$^{\ast}$ &\textbf{79.41$\pm$ 0.24}\\
                           &30 &68.96$\pm$ 0.16  &66.05$\pm$ 0.34  &61.84$\pm$ 0.20 &65.09$\pm$ 0.16 &68.31$\pm$ 0.34$^{\ast}$ &\textbf{73.79$\pm$ 0.31}\\
\hline
\hline
\multirow{4}{*}{MNIST}     &0  &89.12$\pm$ 0.12  &92.05$\pm$ 0.20  &92.19$\pm$ 0.20 &92.84$\pm$ 0.21 &90.68$\pm$ 0.20 &\textbf{93.21$\pm$ 0.20}\\
                           &10 &87.49$\pm$ 0.14  &90.35$\pm$ 0.22  &88.91$\pm$ 0.20 &91.08$\pm$ 0.23 &89.11$\pm$ 0.22 &\textbf{91.79$\pm$ 0.22}\\
                           &20 &85.15$\pm$ 0.14  &84.59$\pm$ 0.26  &83.21$\pm$ 0.20 &84.27$\pm$ 0.28 &86.53$\pm$ 0.26$^{\ast}$ &\textbf{88.41$\pm$ 0.24}\\
                           &30 &80.96$\pm$ 0.16  &78.84$\pm$ 0.34  &76.37$\pm$ 0.20 &78.09$\pm$ 0.16 &80.31$\pm$ 0.34$^{\ast}$ &\textbf{84.38$\pm$ 0.31}\\
\hline
\hline
\multirow{4}{*}{FRGC-2}      &0 &94.39$\pm$  0.17 &95.52$\pm$ 0.21  &93.81$\pm$ 0.21 &\textbf{98.50$\pm$ 0.19} &94.42$\pm$ 0.19 &\textbf{98.50$\pm$ 0.18}\\
                           &10 &89.54$\pm$ 0.18 &88.85$\pm$ 0.25  &85.93$\pm$ 0.21 &90.12$\pm$ 0.25$^{\ast}$ &89.35$\pm$ 0.25 &\textbf{93.61$\pm$ 0.24}\\
                           &20 &83.72$\pm$ 0.21 &80.81$\pm$ 0.31  &77.45$\pm$ 0.21 &81.26$\pm$ 0.33 &84.11$\pm$ 0.29$^{\ast}$ &\textbf{86.34$\pm$ 0.30}\\
                           &30 &76.25$\pm$ 0.25 &73.88$\pm$ 0.38  &69.62$\pm$ 0.21 &74.64$\pm$ 0.36 &76.92$\pm$ 0.32$^{\ast}$ &\textbf{78.89$\pm$ 0.35}\\
\hline
\end{tabular}
\end{center}
\label{tb:real}
\end{table*}

\begin{table*}[t!]
\caption{Comparison of running Time (in seconds) on different datasets.}
\begin{center}
\begin{tabular}{l|c|c|c|c|c}
\hline
Dataset   & NCA  & NCM & LMNN & BML &BNCA  \\
\hline
\hline
Caltech-10   &17.89$\pm$ 0.11  &16.64$\pm$ 0.12  &32.85$\pm$ 0.12 &16.02$\pm$ 0.11 &\textbf{11.72$\pm$ 0.11}\\
\hline
MNIST   &16.60$\pm$ 0.09  &15.95$\pm$ 0.09  &17.95$\pm$ 0.10 &15.51$\pm$ 0.08 &\textbf{10.43$\pm$ 0.08}\\
\hline
FRGC-2   &20.06$\pm$ 0.12  &19.33$\pm$ 0.12  &22.56$\pm$ 0.14 &18.71$\pm$ 0.12 &\textbf{13.50$\pm$ 0.12}\\
\hline
\end{tabular}
\end{center}
\label{tb:time}
\end{table*}

\subsection{Learning from Small Size Training Set}
First we investigate the performance of our method with small sample size on three UCI datasets ("Balance", "Ionosphere", and "Spambase"). In each dataset we randomly sample 3 subsets as training set with the size of 10$\times$C, 20$\times$C and 30$\times$C (C is the number of categaries) respectively, and use an extra subset containing 100 data points as test set.

Table~\ref{tb:uci} gives the classification performance. One can see that when the training set is small, point estimation-based methods tend to be unreliable. With only 10 training samples the standard NCA performs even worse than the baseline approach (PCA) on two of the three datasets tested. By contrast the proposed BNCA performs the best among the compared methods, partly due to the advantage of the Bayesian framework which potentially provides reliable estimation even when the size of the training data is small.

Table~\ref{tb:uci} also shows that with increasing number of training points, the performance of all the methods considered here improves a lot. As expected, when we sample 30 data from each class, the performance gap between the Bayesian approaches and the point estimation based methods (such as LMNN) becomes small. Furthermore, one can see that our BNCA consistently outperforms BML. That illustrates the benefits of graph constraints for metric learning, which allows our method to effectively exploit the local structure of data while BML does not.

\subsection{Learning Under Random Label Noise}\label{sec_mnist_noise}
To test the performance of our method under label noise, we tested our method on several real-world applications, including image classification (on the Caltech-10 dataset \cite{qian2014distance}), digital recognition (on the MNIST \cite{lecun1998gradient}), and face recognition (on the FRGC-2 \cite{phillips2005overview}). The datasets adopted are popular benchmark on each of the task respectively: 1) Caltech-10 is a subset sampled from Caltech-256 image dataset \cite{griffin2007caltech} with 10 most popular categories. The training set contains 300 images (30 from each class) and the test set is another randomly sampled 300 images; 2) The dataset of MNIST we used contains 600 digit images sampled from the full dataset (60 from each class; training/test: 300/300); 3) The dataset of FRGC we used contains 400 face images from 20 subjects (20 images per subject, training/test: 200/200). On all these datasets we inject random label noise on 3 levels ($10\%$, $20\%$, $30\%$) and the test sets are kept clean.

\begin{figure*}[!t]
\centering
\subfigure[]{\includegraphics[width=0.45\textwidth]{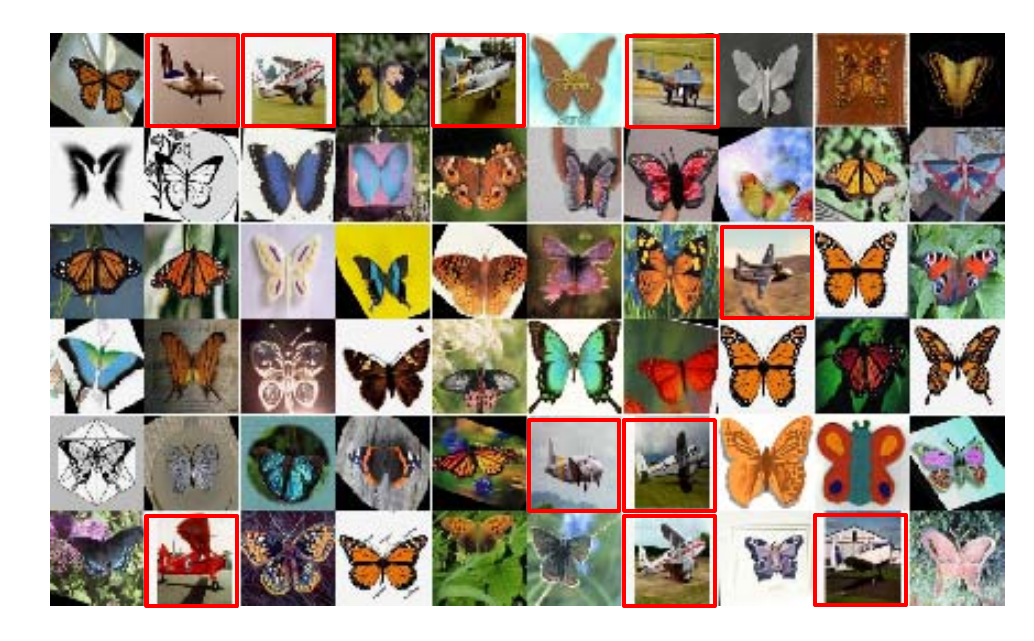}
\label{fig:cal_10a}}
\subfigure[]{\includegraphics[width=0.45\textwidth]{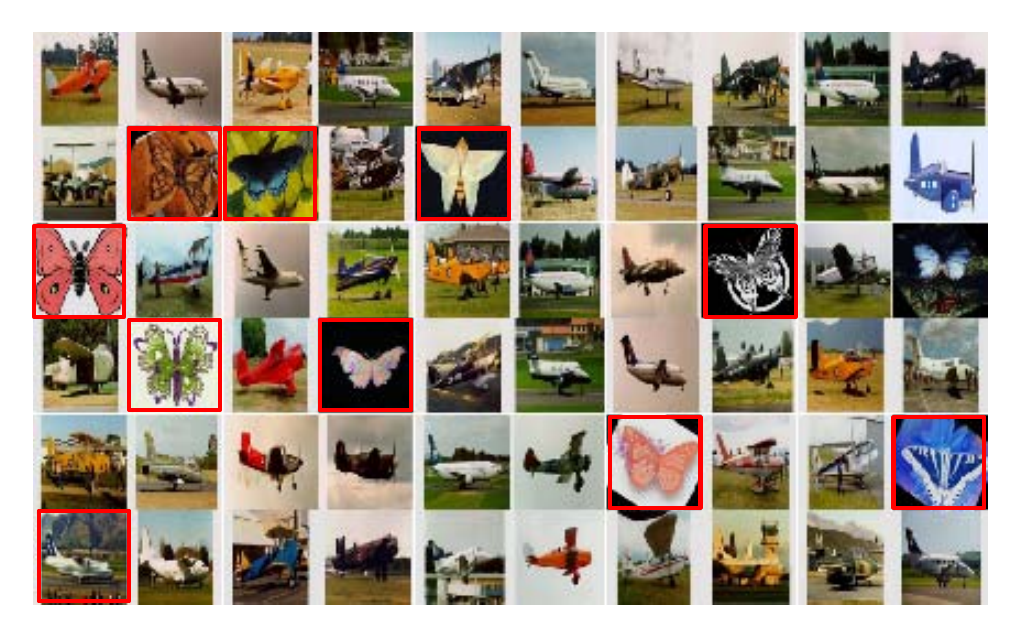}
\label{fig:cal_10b}}
\caption{Illustration of training images injected with random label noise from the category of ``plane" (a) and ``butterfly" (b) in the Caltech-10 dataset, where images with inaccurate labels are marked with a red square.}
\label{fig:cal10}
\end{figure*}

Fig.~\ref{fig:cal10} illustrates some of the noisy data of Caltech-10. One can see that in each category there exist some portion of images that are not belonging to this category, possibly due to the errors introduced in the labelling procedure, and very few work investigate the consequence of this.

Table~\ref{tb:real} shows how the metric learning algorithms perform under random label noise. Label noise could mislead metric learning algorithms in a way that it pulls data from different class together while keeps those from the same class away. When there is no label noise, almost all metric learning methods help to make an improvement in accuracy. However, it can be seen that the performance of all the methods declines with the increasing of noise level. Particularly, as the noise level increases to $30\%$ some of the metric learning methods do not work (such as NCA, NCM and LMNN) in the sense that they even perform worse than the original baseline PCA. Our BNCA works significantly better than traditional metric learning methods even under this challenging case - even when the noise level reaches 30\%, the p-value is smaller than 0.001 when comparing our method with the second best performer in terms of accuracy.

Table~\ref{tb:time} compares the corresponding running time of the methods in Table~\ref{tb:real} under the situation of no label noise, with our (unoptimized Matlab) implementation. The table shows that on the average the proposed BNCA runs more 61.2\% faster than the NCA algorithm and more than 41.0\% faster than the BML method. This is consistent with our analysis described section.~\ref{sec_anltime}. That is, in each iteration of variational inference, the introduction of fixed curvature Bohning bound effectively avoids computing the Hessian matrix, resulting in significant reduction of running time.

\subsection{Predictive Performance under Difficult Conditions}
\begin{figure*}[!t]
\centering
\subfigure[]{\includegraphics[width=0.45\textwidth]{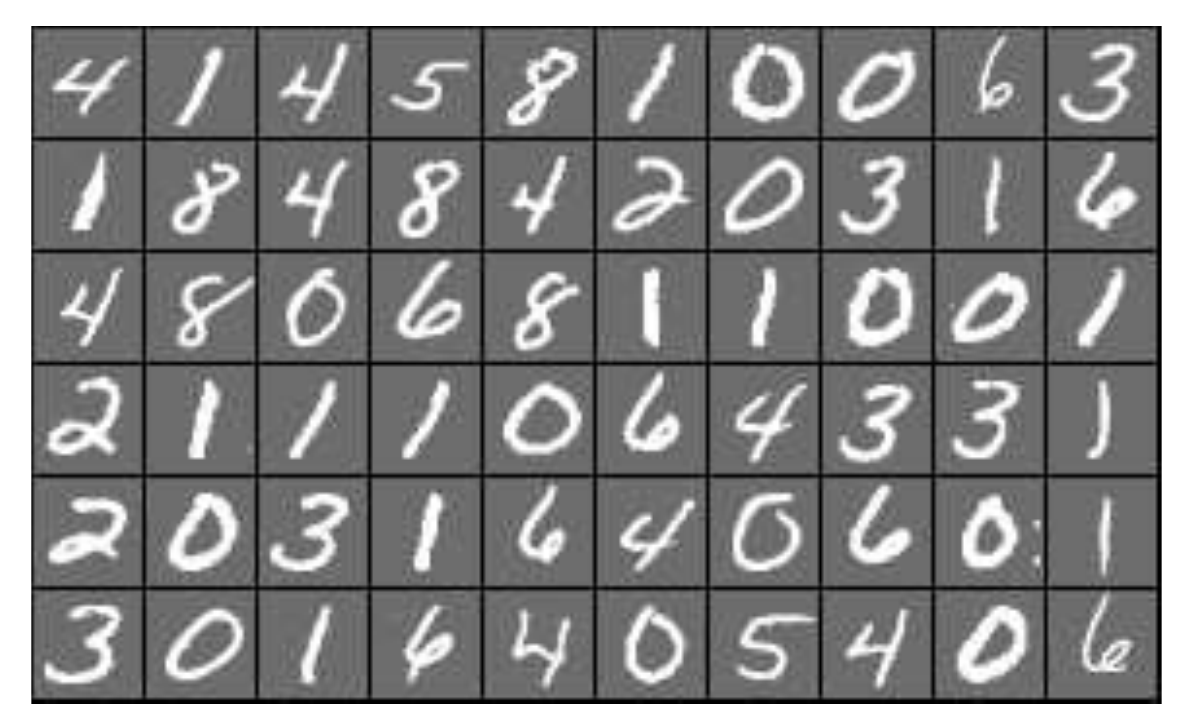}
\label{fig:mnist_normal}}
\subfigure[]{\includegraphics[width=0.45\textwidth]{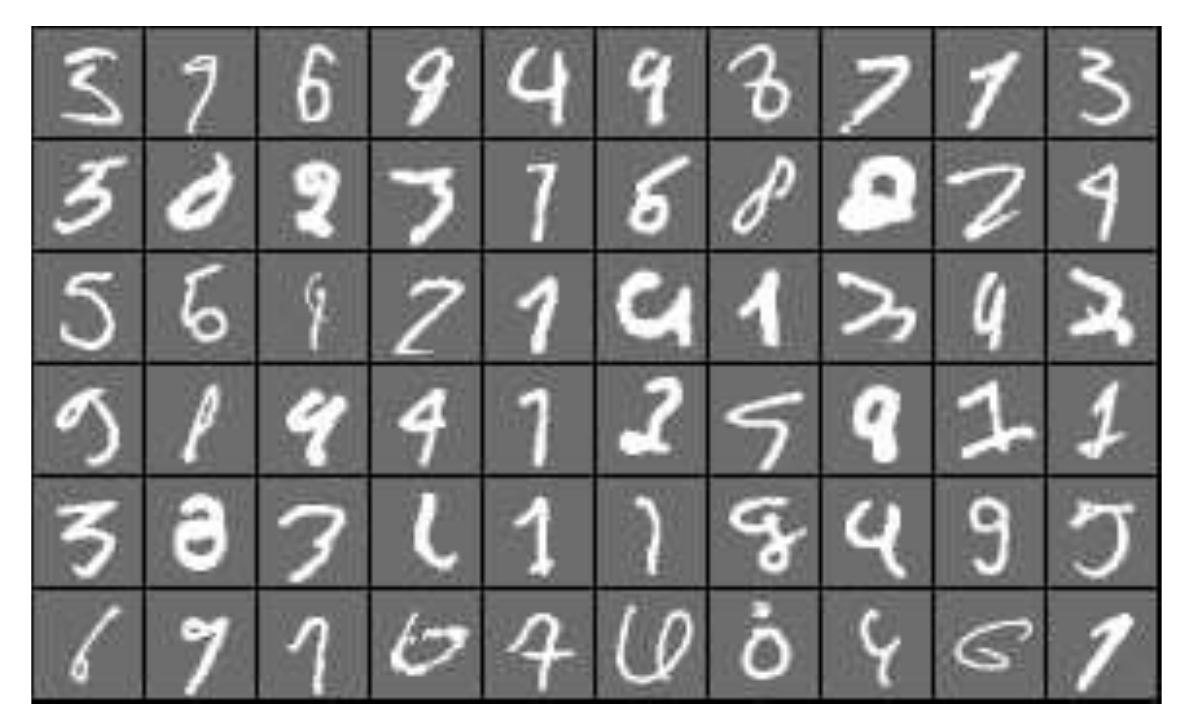}
\label{fig:mnist_hard}}
\caption{Illustration of the data from the MNIST datasets: (a) "normal" data (b) "difficult" data.}
\label{fig:mnist}
\end{figure*}

\begin{figure*}[!t]
\centering
\subfigure[]{\includegraphics[width=0.95\textwidth]{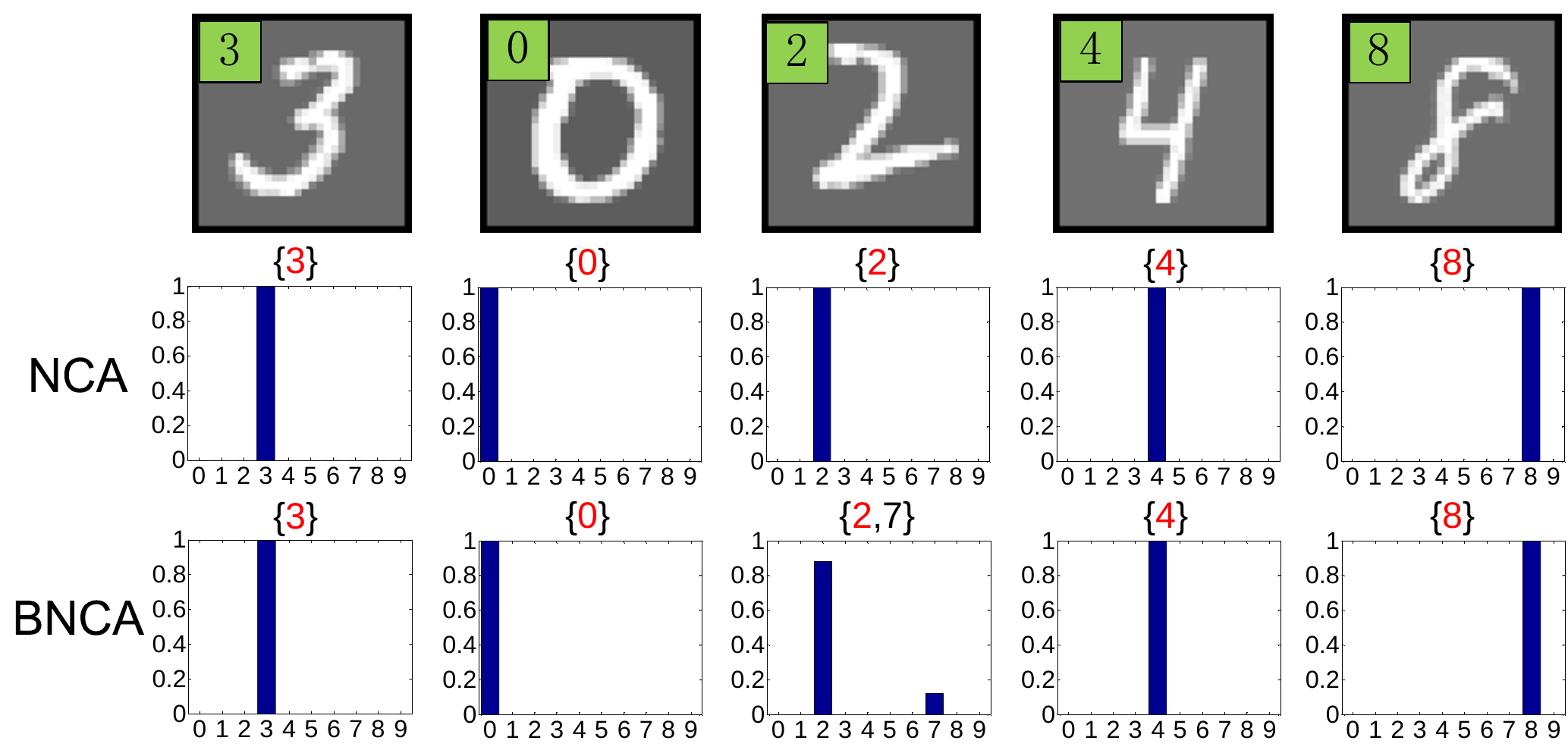}
\label{fig:uncertainty_normal}}
\subfigure[]{\includegraphics[width=0.95\textwidth]{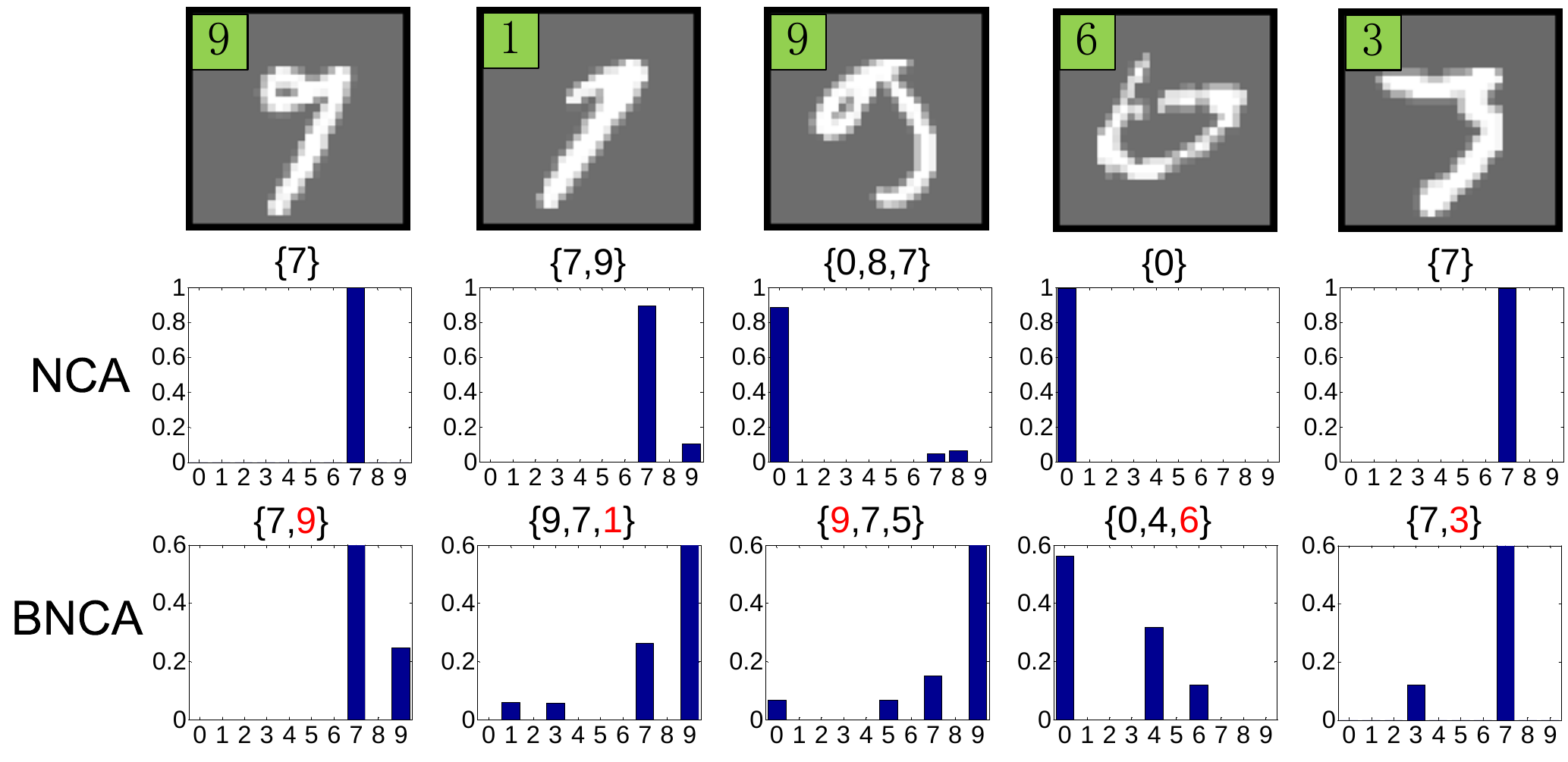}
\label{fig:uncertainty_hard}}
\caption{Visualization of predictive probability: $P(y_i=k|x_i,Y_{N_i},X_{N_i},A)$ and $P(y_i=k|x_i,Y_{N_i}, X_{N_i})$. (a) is the result on normal test data that the distribution of both NCA and BNCA has a single peak probability mass. (b) is on difficult test data that NCA still has a single peak while BNCA assigns probability to several possible candidates. The digit in green square is the groundtruth label and the digits in braces is the ranking list of predictions.}
\label{fig:uncertainty}
\end{figure*}

\begin{figure}
\centering
\includegraphics[width=1\linewidth]{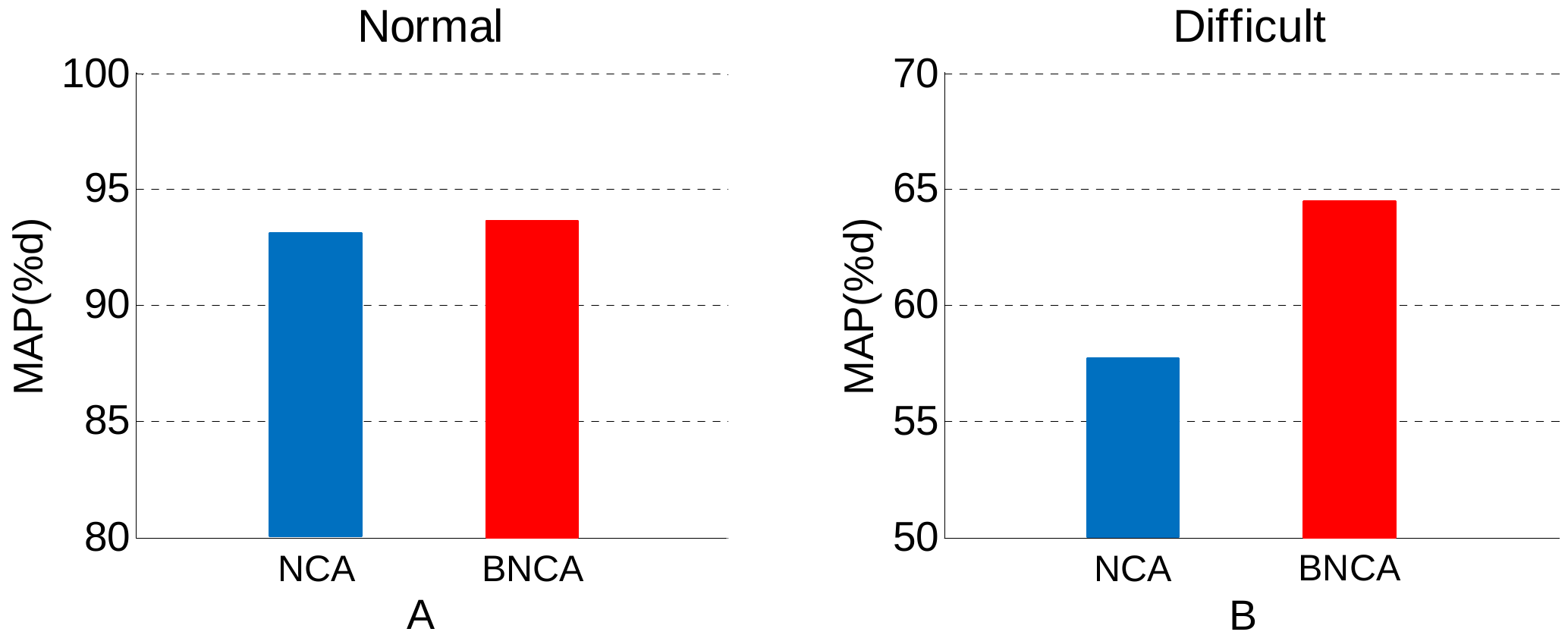}
\caption{Comparison of Mean Average Precision ($\%$) of NCA and BNCA on the "normal" test set (a) and on the "difficult" test set (b).}
\label{fig:hard_hist}
\end{figure}

In the previous two sections we have shown the benefits of the proposed BNCA method that learns either from a small number of training examples or from examples with label noise. In this section, we investigate empirically the robustness performance of the BNCA method under difficult conditions by comparing it with the baseline NCA method. The motivation for this is that since the difficult samples are commonly those lying either in the uncertain region, or those lying far away from the normal distribution, making prediction under these conditions would impose great challenge for a traditional DML method based on point estimation, due to its lack of accounting for parameter uncertainty.

We conducted this series of experiments using MNIST \cite{lecun1998gradient}. Firstly 300 "normal" data points are sampled (by "normal" we mean that those digital images are not difficult for a human to recognize), and are used to train two models, i.e., a NCA and a BNCA model. For test, we collect two different test sets: One is normal while the other is most difficult in the sense that all digital images in this set are hard to recognize even by human. Since it is both time-consuming and error prone to select those difficult samples manually, we adopt one state of the art model on the MNIST dataset, i.e, the C-SVDDNet \cite{wang2014unsupervised}, as the expert to choose samples, and those samples close to the decision boundary of C-SVDDNet would be regarded as difficult samples， otherwise as normal samples. In this way, we collect 300 random normal samples and 300 most difficult samples respectively as test sets. There is no overlapping between these two test sets. Some of the samples are illustrated in Fig.~\ref{fig:mnist}. To evaluate the performance of NCA and Bayesian NCA in the two cases, we compute the predictive probability $P(y_i|x_i,Y_{N_i},X_{N_i},A)$ (using eq.~(\ref{eq_likli1})) and $P(y_i|x_i,Y_{N_i},X_{N_i})$ (using eq.~(\ref{eq_uncertainty_mcmc})) on the test data.

Fig.~\ref{fig:uncertainty_normal} visualizes the probability mass assigned to the normal test samples by the two models, respectively. We can see that both NCA and BNCA have a single peak probability mass, indicating that both of them are quite certain about their predictions. However, on the difficult sets their behaviors are largely  different. Fig.~\ref{fig:uncertainty_hard} gives the results on this harder test set, and the ranking list of predictions according to their assigned probability.

It is obvious that the NCA is overconfident in its prediction. For example, the leftmost column of Fig.~\ref{fig:uncertainty_hard} shows that NCA incorrectly classifies the image of "9" as "7" with a high predictive probability of over 0.99 (those predictions with posterior mass less than 0.01 are canceled), indicating that this type of approximation to the posterior with a point mass is inadequate. On the other hand, the predictions of the BNCA are more moderate. One can see that although the true labels may not be ranked the highest, they are correctly appeared among the first few high ranking candidates. Under the previous example, although there is over 60\% probability is assigned to the digital of "7" by our method, a significantly higher amount of predictive mass ($\ge$ 25\%) than that of NCA is correctly assigned to the number of "9". This reveals that under difficult conditions BNCA provides a much better approximation to the posterior than the point estimation method of NCA, by considering the uncertainty of parameters.

More precisely, we compare the performance of NCA and BNCA on the two test sets. For this a new measurement, i.e., modified mean average precision (MAP), is introduced as our performance metric, which is defined as
\begin{equation}
\begin{split}
MAP &= \sum_{i=1}^N\sum_{k=1}^Kp_i(k)\cdot 1\{y_i = k\} \cdot 1\{p(y_i = k|x_i) > \tau \}
\end{split}\label{eq_map}
\end{equation}
where $p_i(k)$ is the precision at cut-off $k$ in $i$'s ranking list, and $\tau$ is a truncating threshold. The threshold is usually set to be a very small number (e.g., 0.01), since if the corresponding response $p(y_i = k|x_i)$ of the model to the input $x_i$ is too small, there is no point to count it in performance measuring.

Fig.~\ref{fig:hard_hist} gives the results. One can see that while on the normal test data, the MAP accuracy of BNCA is slightly better than that of NCA (about $1.0\%$ higher), the BNCA significantly outperforms NCA by more than $5.0\%$ on the difficult set. This reveals that taking the prediction uncertainty into account indeed helps to improve the mean average precision. Also note that since the pairwise constrained BML method \cite{Yang07} only estimates whether a date pair belongs to the same class or not, while not being able to give the predictive distribution $p(y_i|x_i)$, it is not included for comparison here.

\subsection{Discussions}

\subsubsection{Robustness against Overfitting}\label{sec_stable}
\begin{figure*}[!t]
\centering
\subfigure[]{\includegraphics[width=0.32\textwidth]{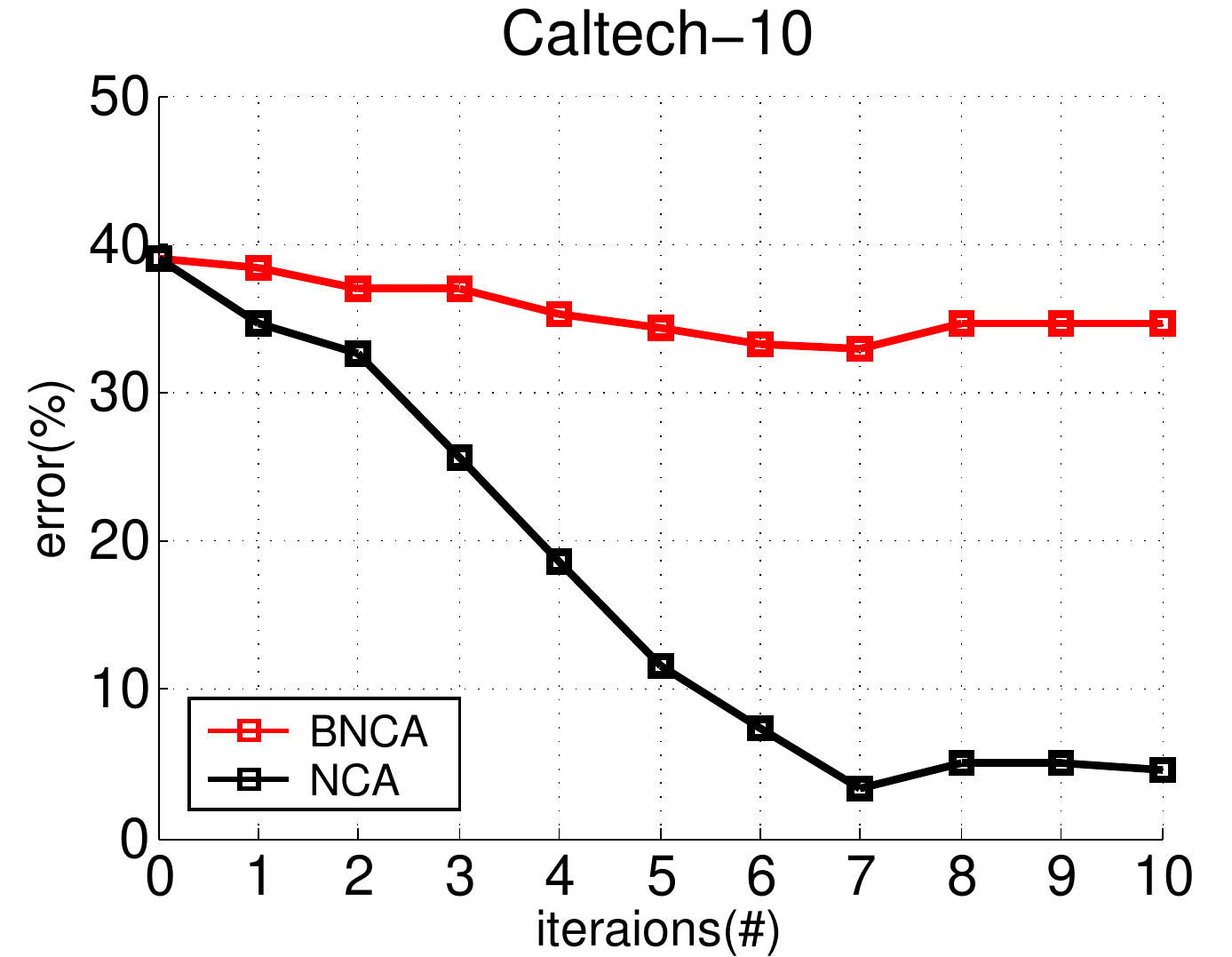}
\label{fig:itercal_tr}}
\subfigure[]{\includegraphics[width=0.32\textwidth]{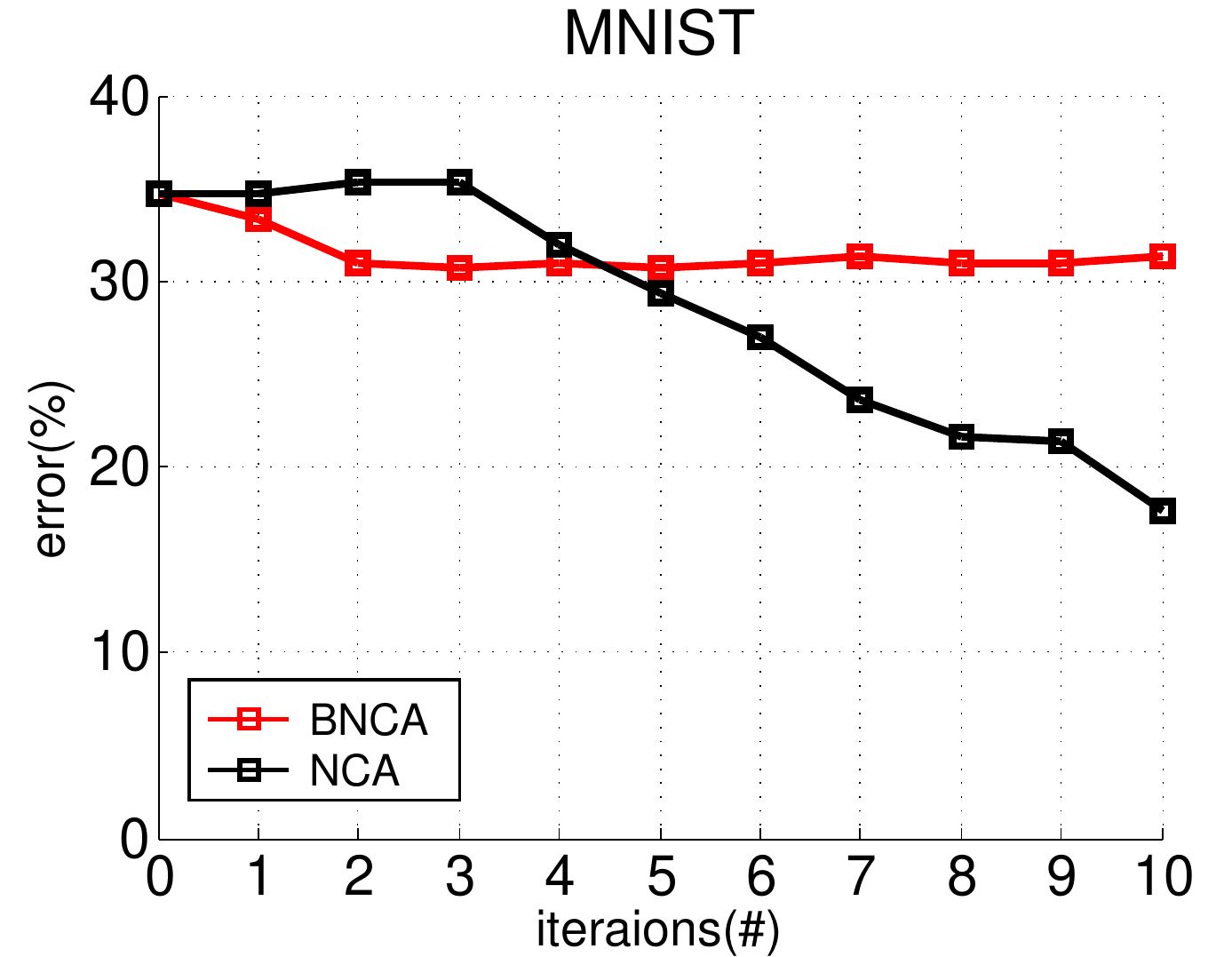}
\label{fig:itermnist_tr}}
\subfigure[]{\includegraphics[width=0.32\textwidth]{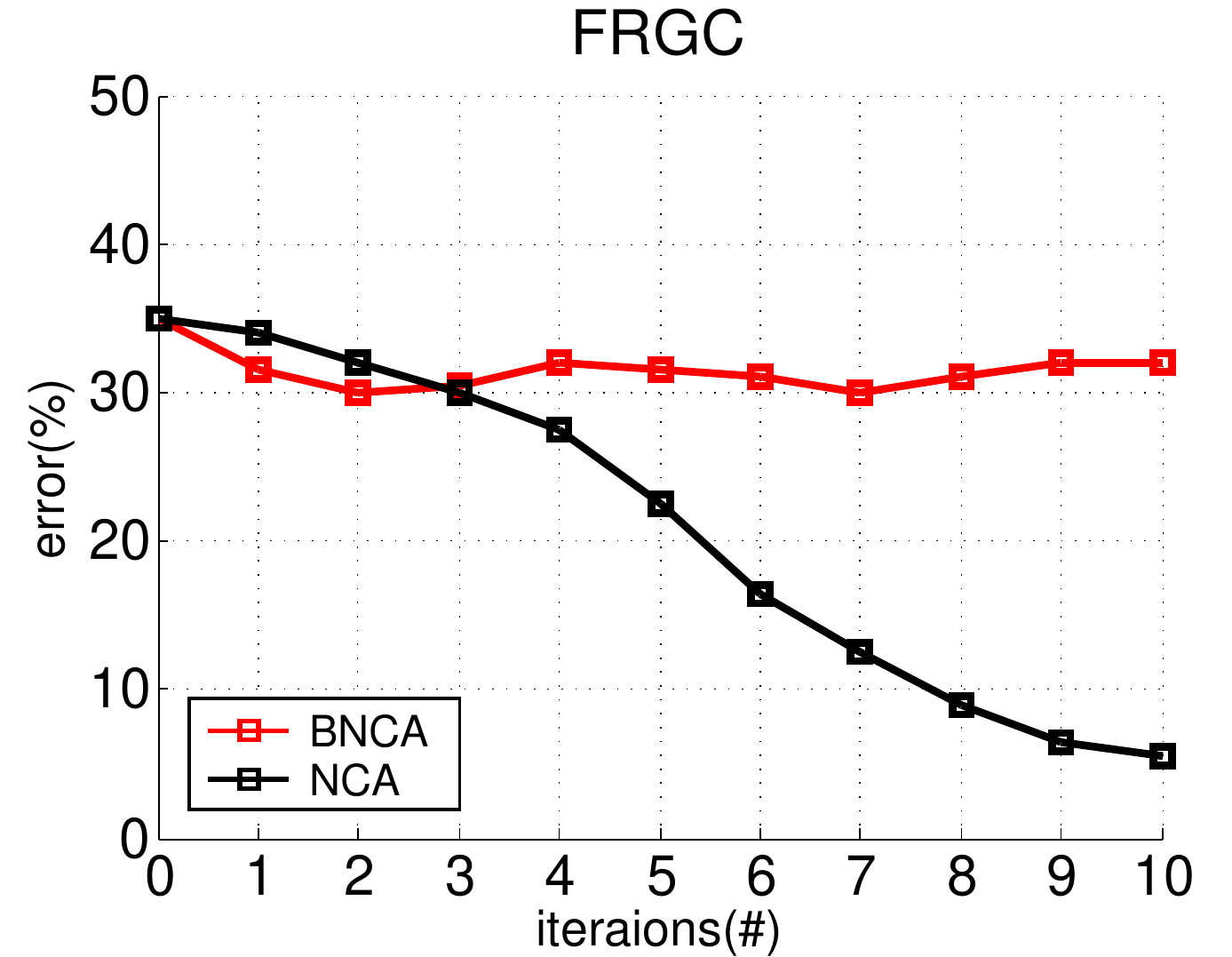}
\label{fig:iterfrgc_tr}}
\subfigure[]{\includegraphics[width=0.32\textwidth]{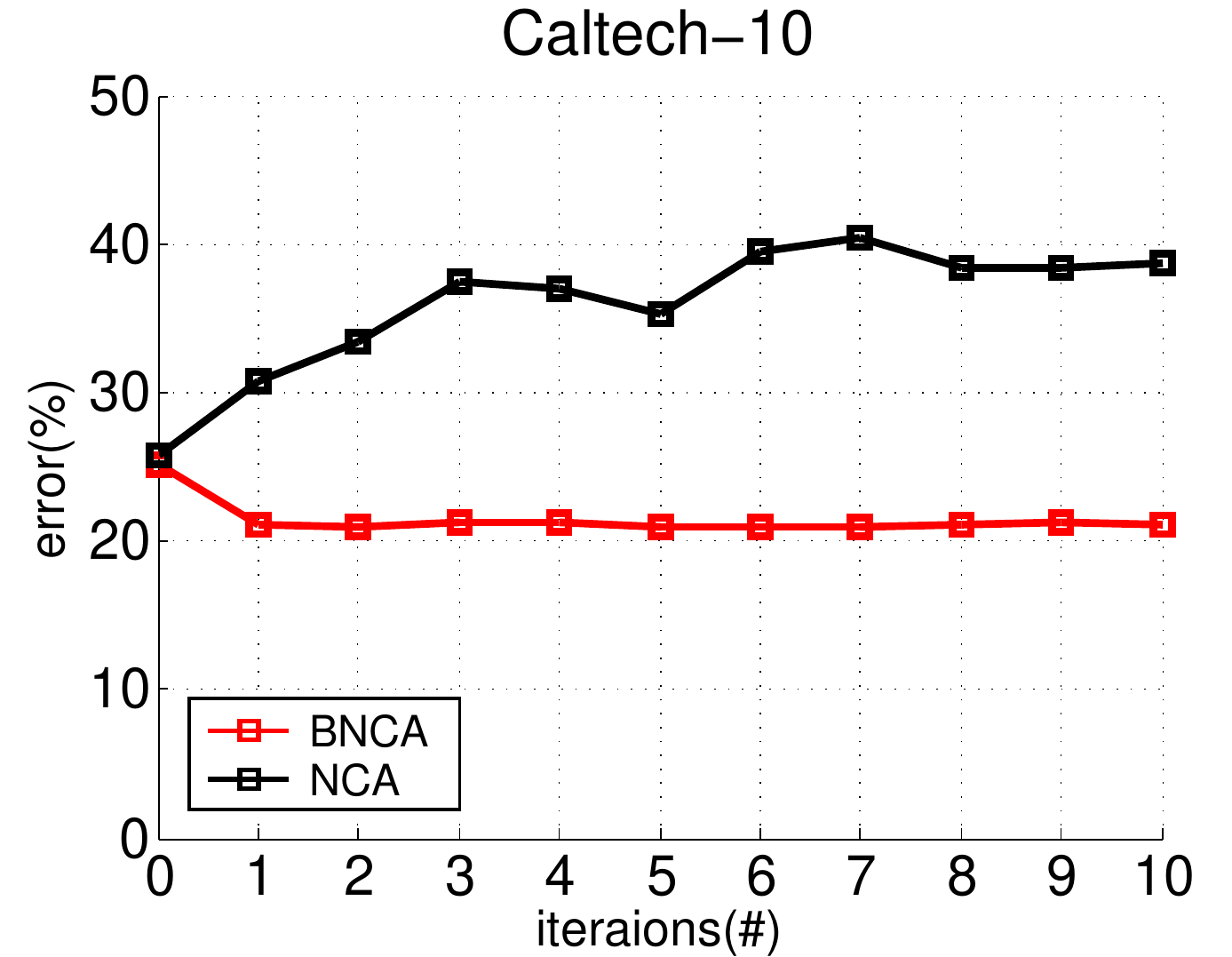}
\label{fig:itercal}}
\subfigure[]{\includegraphics[width=0.32\textwidth]{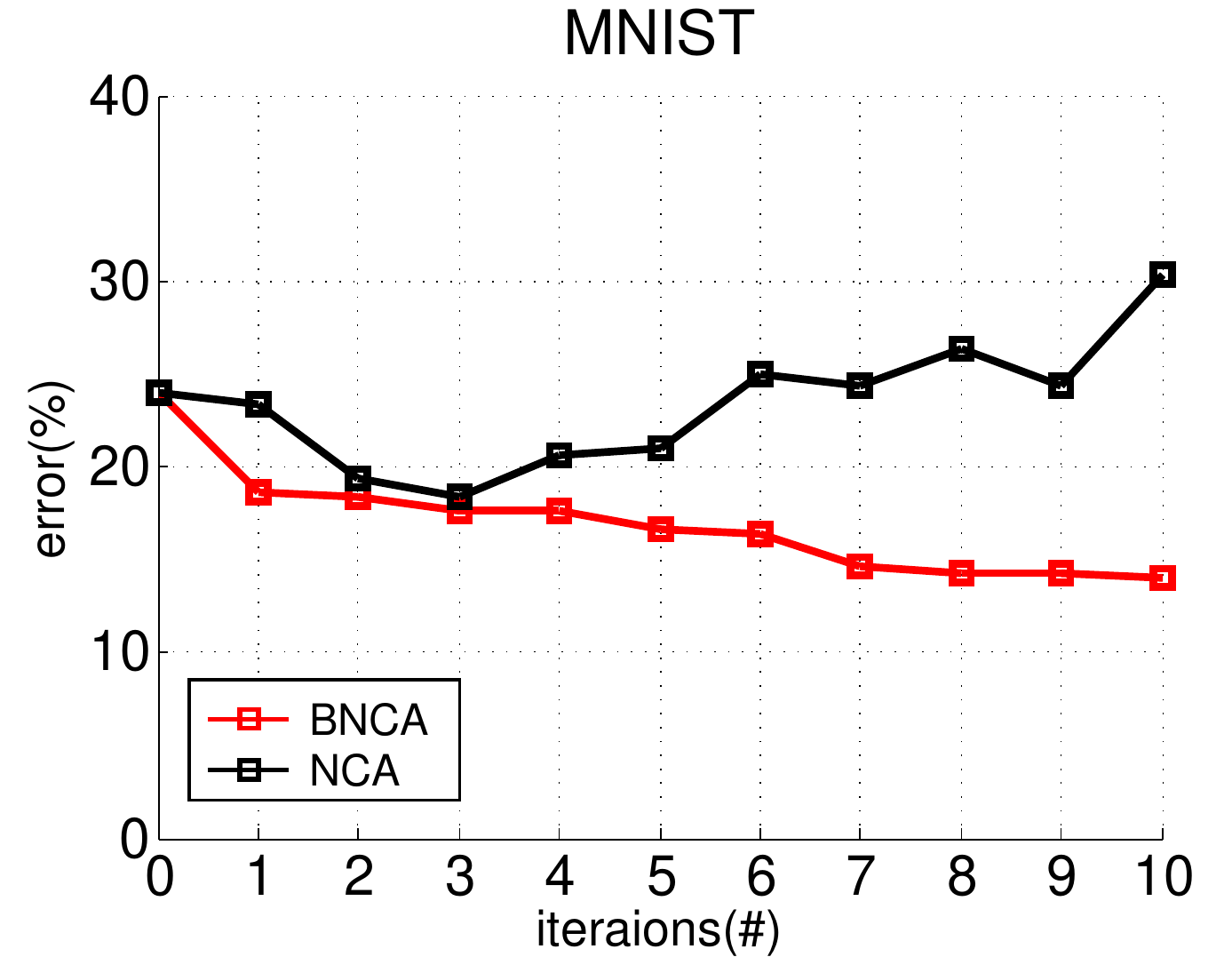}
\label{fig:itermnist}}
\subfigure[]{\includegraphics[width=0.32\textwidth]{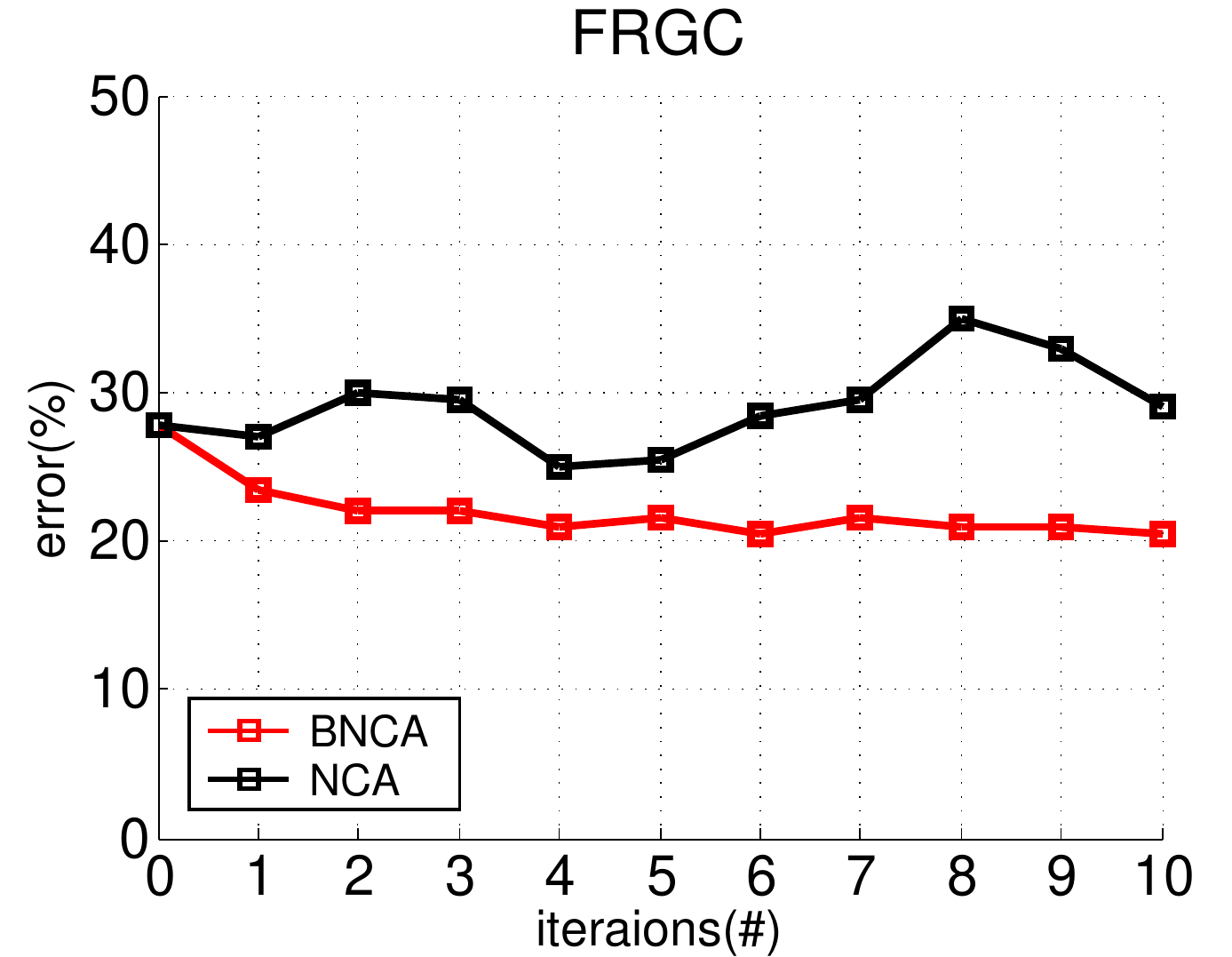}
\label{fig:iterfrgc}}
\caption{Learning curves of NCA and BNCA. The classifier is KNN and the noise level is at $30\%$. (a), (b), (c) are with the training set while (d), (e), (f) are with the test set}
\label{fig:iter}
\end{figure*}
To further investigate the behavior of the proposed BNCA method, we plot in Fig.~\ref{fig:iter} the learning curves of both BNCA and NCA as the function of the number of iterations. Three datasets (Caltech-10, MNIST, and FRGC) are used for this, with the same experimental setting as before, and for each dataset there are 30.0\% random label noise injected. For NCA training, we used the conjugate gradient method \cite{shewchuk1994introduction}, which seeks the steepest gradient direction with proper step size in each training step. The figure shows that with the iterations going, on all of the three datasets the training errors of NCA keep decreases but their test errors tend to rise at the same time, indicating that the method is easy to be overfitting under the condition of label noise. Although some empirical tricks such as early stopping can be adopted, the figure clearly shows that this is not an issue for our Bayesian extension to the NCA. Actually the figure reveals that it only takes a few iterations before the learning converges.

\subsubsection{The Importance of Prior in Metric Learning}
\begin{figure}[t!]
\centering
\includegraphics[width=0.9\linewidth]{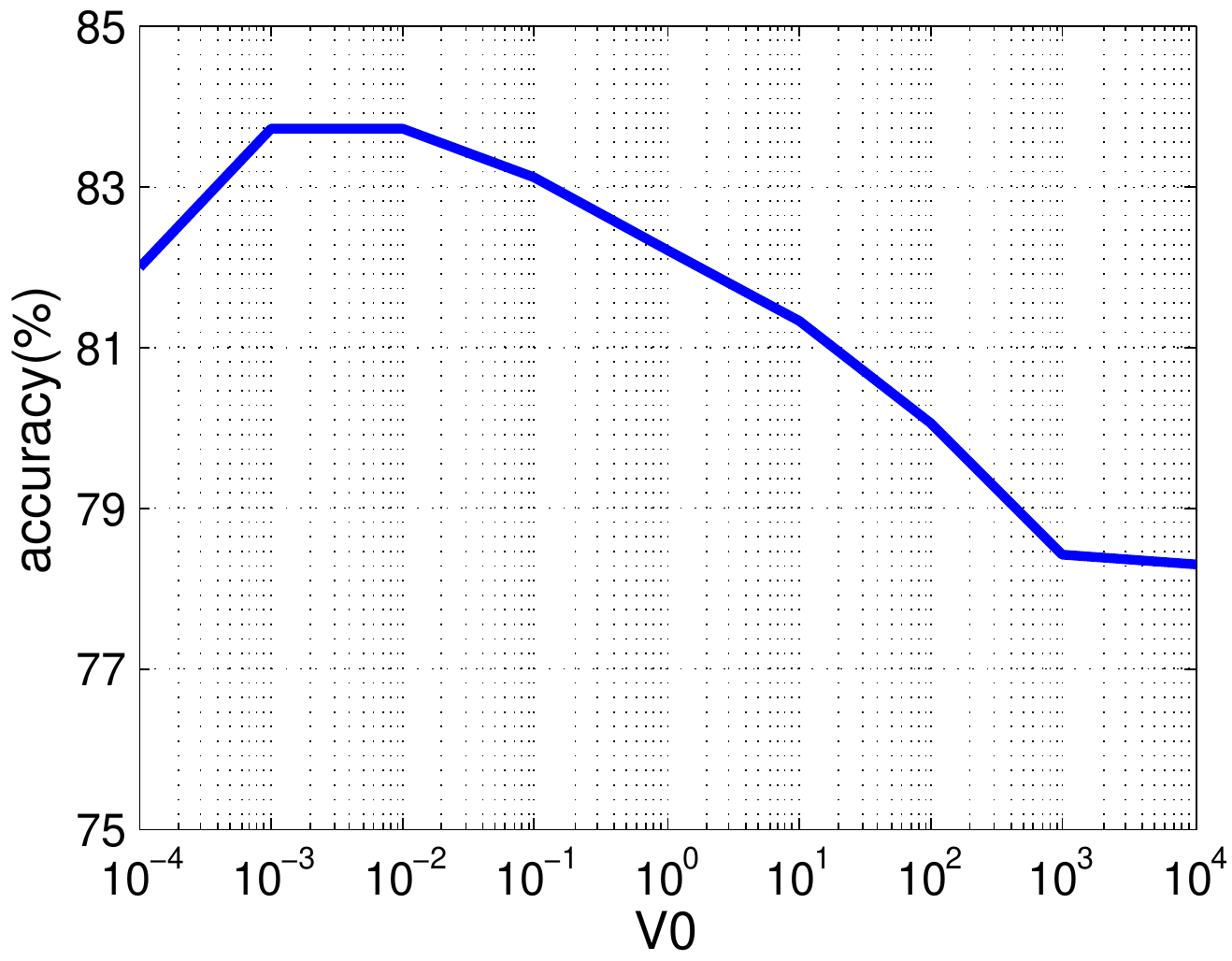}
\caption{The influence of prior on the performance on the Caltech-10 dataset.}
\label{fig:v0}
\end{figure}

As described in Section~\ref{sec_proposedmethod}, our method  explores the usefulness of prior within a Bayesian framework, whose parameters are efficiently estimated through variational inference. It would be interesting to investigate in more details on the role played by the prior. For this we conducted a series of experiments on the Caltech-10 dataset by varying the value of $V_0$ ($\sigma I$) from $10^{-4}$ to $10^4$ but keeping the mean value $m_0$ fixed at the same time. Note that a large value of $V_0$ indicates that the prior tends to be more noninformative (i.e., higher uncertain) about the $\gamma$ value. Fig.~\ref{fig:v0} shows how the performance changes as a function of the degree of uncertainty in prior. One can see that the prior is beneficial when the value of $V_0$ in the range between $10^{-4}$ and $10^{0}$, but further increasing this tends to be useless as the prior becomes more noninformative.

We note that as it is generally difficult to estimate the full matrix $A$ (c.f., eq.~(\ref{eq_propri})), most traditional metric learning methods exploit various low-rank approximation \cite{goldberger2004neighbourhood} \cite{mensink2012metric} to it without using any prior. This may encounter difficulty especially when the size of samples is small or when the labels are noisy, and Bayesian learning provides a unique way to address this issue by incorporating the prior information.

\section{Conclusion}\label{sec_conclude}
We present a new Bayesian metric learning method---Bayesian Neighbourhood component analysis (BNCA) that effectively improves the performance of KNN classifier under the condition of small sample size and/or when data labels are noisy. The method is based on the classical NCA method with point estimation, and for the first time extends it under the Bayesian framework. The major advantages of BNCA over NCA in distance metric learning are three folds: 1) it is easy to train without worrying about overfitting; 2) it performs more robust compared to NCA under difficult conditions; 3) it naturally handles label noise by reducing the influence of data points with possible labelling errors. In addition, to improve the efficiency of Bayesian learning, we introduce a new variational lower bound of the log-likelihood of the objective. Extensive experiments conducted on several challenging real-world applications show that the performance of the proposed BNCA method significantly improves upon the baseline NCA method and it outperforms several other state of the art distance metric learning methods as well. We are currently investigating more applications of the proposed BNCA method, such as image retrieval with model uncertainty.
\bibliographystyle{IEEEtran}
\bibliography{refbibtex}

\begin{thebibliography}{10}
\providecommand{\url}[1]{#1}
\csname url@samestyle\endcsname
\providecommand{\newblock}{\relax}
\providecommand{\bibinfo}[2]{#2}
\providecommand{\BIBentrySTDinterwordspacing}{\spaceskip=0pt\relax}
\providecommand{\BIBentryALTinterwordstretchfactor}{4}
\providecommand{\BIBentryALTinterwordspacing}{\spaceskip=\fontdimen2\font plus
\BIBentryALTinterwordstretchfactor\fontdimen3\font minus
  \fontdimen4\font\relax}
\providecommand{\BIBforeignlanguage}[2]{{%
\expandafter\ifx\csname l@#1\endcsname\relax
\typeout{** WARNING: IEEEtran.bst: No hyphenation pattern has been}%
\typeout{** loaded for the language `#1'. Using the pattern for}%
\typeout{** the default language instead.}%
\else
\language=\csname l@#1\endcsname
\fi
#2}}
\providecommand{\BIBdecl}{\relax}
\BIBdecl

\bibitem{mensink2012metric}
T.~Mensink, J.~Verbeek, F.~Perronnin, and G.~Csurka, ``Metric learning for
  large scale image classification: Generalizing to new classes at near-zero
  cost,'' in \emph{Computer Vision--ECCV 2012}.\hskip 1em plus 0.5em minus
  0.4em\relax Springer, 2012, pp. 488--501.

\bibitem{hoi2006learning}
S.~C. Hoi, W.~Liu, M.~R. Lyu, and W.-Y. Ma, ``Learning distance metrics with
  contextual constraints for image retrieval,'' in \emph{Computer Vision and
  Pattern Recognition, 2006 IEEE Computer Society Conference on}, vol.~2.\hskip
  1em plus 0.5em minus 0.4em\relax IEEE, 2006, pp. 2072--2078.

\bibitem{li2015ordinal}
C.~Li, Q.~Liu, J.~Liu, and H.~Lu, ``Ordinal distance metric learning for image
  ranking,'' \emph{Neural Networks and Learning Systems, IEEE Transactions on},
  vol.~26, no.~7, pp. 1551--1559, 2015.

\bibitem{guillaumin2010multiple}
M.~Guillaumin, J.~Verbeek, and C.~Schmid, ``Multiple instance metric learning
  from automatically labeled bags of faces,'' in \emph{Computer Vision--ECCV
  2010}.\hskip 1em plus 0.5em minus 0.4em\relax Springer, 2010, pp. 634--647.

\bibitem{lu2014neighborhood}
J.~Lu, X.~Zhou, Y.-P. Tan, Y.~Shang, and J.~Zhou, ``Neighborhood repulsed
  metric learning for kinship verification,'' \emph{Pattern Analysis and
  Machine Intelligence, IEEE Transactions on}, vol.~36, no.~2, pp. 331--345,
  2014.

\bibitem{ye2007adaptive}
J.~Ye, Z.~Zhao, and H.~Liu, ``Adaptive distance metric learning for
  clustering,'' in \emph{Computer Vision and Pattern Recognition, 2007.
  CVPR'07. IEEE Conference on}.\hskip 1em plus 0.5em minus 0.4em\relax IEEE,
  2007, pp. 1--7.

\bibitem{dikmen2011pedestrian}
M.~Dikmen, E.~Akbas, T.~S. Huang, and N.~Ahuja, ``Pedestrian recognition with a
  learned metric,'' in \emph{Computer Vision--ACCV 2010}.\hskip 1em plus 0.5em
  minus 0.4em\relax Springer, 2011, pp. 501--512.

\bibitem{guillaumin2009you}
M.~Guillaumin, J.~Verbeek, and C.~Schmid, ``Is that you? metric learning
  approaches for face identification,'' in \emph{Computer Vision, 2009 IEEE
  12th International Conference on}.\hskip 1em plus 0.5em minus 0.4em\relax
  IEEE, 2009, pp. 498--505.

\bibitem{guillaumin2009tagprop}
M.~Guillaumin, T.~Mensink, J.~Verbeek, and C.~Schmid, ``Tagprop: Discriminative
  metric learning in nearest neighbor models for image auto-annotation,'' in
  \emph{Computer Vision, 2009 IEEE 12th International Conference on}.\hskip 1em
  plus 0.5em minus 0.4em\relax IEEE, 2009, pp. 309--316.

\bibitem{koestinger2012large}
M.~Koestinger, M.~Hirzer, P.~Wohlhart, P.~M. Roth, and H.~Bischof, ``Large
  scale metric learning from equivalence constraints,'' in \emph{Computer
  Vision and Pattern Recognition (CVPR), 2012 IEEE Conference on}.\hskip 1em
  plus 0.5em minus 0.4em\relax IEEE, 2012, pp. 2288--2295.

\bibitem{bian2012constrained}
W.~Bian and D.~Tao, ``Constrained empirical risk minimization framework for
  distance metric learning,'' \emph{Neural Networks and Learning Systems, IEEE
  Transactions on}, vol.~23, no.~8, pp. 1194--1205, 2012.

\bibitem{fouad2013incorporating}
S.~Fouad, P.~Tino, S.~Raychaudhury, and P.~Schneider, ``Incorporating
  privileged information through metric learning,'' \emph{Neural Networks and
  Learning Systems, IEEE Transactions on}, vol.~24, no.~7, pp. 1086--1098,
  2013.

\bibitem{shen2014efficient}
C.~Shen, J.~Kim, F.~Liu, L.~Wang, and A.~Van Den~Hengel, ``Efficient dual
  approach to distance metric learning,'' \emph{Neural Networks and Learning
  Systems, IEEE Transactions on}, vol.~25, no.~2, pp. 394--406, 2014.

\bibitem{li2015distributed}
J.~Li, X.~Lin, X.~Rui, Y.~Rui, and D.~Tao, ``A distributed approach toward
  discriminative distance metric learning,'' \emph{Neural Networks and Learning
  Systems, IEEE Transactions on}, vol.~26, no.~9, pp. 2111--2122, 2015.

\bibitem{Yang07}
L.~Yang, R.~Jin, and R.~Sukthankar, ``Bayesian active distance metric
  learning,'' in \emph{Proceedings of the Twenty-Third Conference Annual
  Conference on Uncertainty in Artificial Intelligence (UAI-07)}.\hskip 1em
  plus 0.5em minus 0.4em\relax Corvallis, Oregon: AUAI Press, 2007, pp.
  442--449.

\bibitem{goldberger2004neighbourhood}
J.~Goldberger, G.~E. Hinton, S.~T. Roweis, and R.~Salakhutdinov,
  ``Neighbourhood components analysis,'' in \emph{Advances in neural
  information processing systems}, 2004, pp. 513--520.

\bibitem{weinberger2005distance}
K.~Q. Weinberger, J.~Blitzer, and L.~K. Saul, ``Distance metric learning for
  large margin nearest neighbor classification,'' in \emph{Advances in neural
  information processing systems}, 2005, pp. 1473--1480.

\bibitem{wang2014robust}
D.~Wang and X.~Tan, ``Robust distance metric learning in the presence of label
  noise,'' in \emph{Twenty-Eighth AAAI Conference on Artificial Intelligence},
  2014.

\bibitem{kearns1994toward}
M.~J. Kearns, R.~E. Schapire, and L.~M. Sellie, ``Toward efficient agnostic
  learning,'' \emph{Machine Learning}, vol.~17, no. 2-3, pp. 115--141, 1994.

\bibitem{lawrence2001estimating}
N.~D. Lawrence and B.~Sch{\"o}lkopf, ``Estimating a kernel fisher discriminant
  in the presence of label noise,'' in \emph{ICML}.\hskip 1em plus 0.5em minus
  0.4em\relax Citeseer, 2001, pp. 306--313.

\bibitem{cantador2005boosting}
I.~Cantador and J.~R. Dorronsoro, ``Boosting parallel perceptrons for label
  noise reduction in classification problems,'' in \emph{Artificial
  Intelligence and Knowledge Engineering Applications: A Bioinspired
  Approach}.\hskip 1em plus 0.5em minus 0.4em\relax Springer, 2005, pp.
  586--593.

\bibitem{van2010novel}
J.~Van~Hulse, T.~M. Khoshgoftaar, and A.~Napolitano, ``A novel noise filtering
  algorithm for imbalanced data,'' in \emph{Machine Learning and Applications
  (ICMLA), 2010 Ninth International Conference on}.\hskip 1em plus 0.5em minus
  0.4em\relax IEEE, 2010, pp. 9--14.

\bibitem{leung2011handling}
T.~Leung, Y.~Song, and J.~Zhang, ``Handling label noise in video classification
  via multiple instance learning,'' in \emph{Computer Vision (ICCV), 2011 IEEE
  International Conference on}.\hskip 1em plus 0.5em minus 0.4em\relax IEEE,
  2011, pp. 2056--2063.

\bibitem{fefilatyev2012label}
S.~Fefilatyev, M.~Shreve, K.~Kramer, L.~Hall, D.~Goldgof, R.~Kasturi, K.~Daly,
  A.~Remsen, and H.~Bunke, ``Label-noise reduction with support vector
  machines,'' in \emph{Pattern Recognition (ICPR), 2012 21st International
  Conference on}.\hskip 1em plus 0.5em minus 0.4em\relax IEEE, 2012, pp.
  3504--3508.

\bibitem{frenay2013classification}
B.~Fr{\'e}nay and M.~Verleysen, ``Classification in the presence of label
  noise: a survey,'' 2013.

\bibitem{murphy2012machine}
K.~P. Murphy, \emph{Machine learning: a probabilistic perspective}.\hskip 1em
  plus 0.5em minus 0.4em\relax MIT press, 2012.

\bibitem{andrieu2003introduction}
C.~Andrieu, N.~De~Freitas, A.~Doucet, and M.~I. Jordan, ``An introduction to
  mcmc for machine learning,'' \emph{Machine learning}, vol.~50, no. 1-2, pp.
  5--43, 2003.

\bibitem{bottou2010large}
L.~Bottou, ``Large-scale machine learning with stochastic gradient descent,''
  in \emph{Proceedings of COMPSTAT'2010}.\hskip 1em plus 0.5em minus
  0.4em\relax Springer, 2010, pp. 177--186.

\bibitem{shewchuk1994introduction}
J.~R. Shewchuk, ``An introduction to the conjugate gradient method without the
  agonizing pain,'' 1994.

\bibitem{wold1987principal}
S.~Wold, K.~Esbensen, and P.~Geladi, ``Principal component analysis,''
  \emph{Chemometrics and intelligent laboratory systems}, vol.~2, no. 1-3, pp.
  37--52, 1987.

\bibitem{qian2014distance}
Q.~Qian, J.~Hu, R.~Jin, J.~Pei, and S.~Zhu, ``Distance metric learning using
  dropout: a structured regularization approach,'' in \emph{Proceedings of the
  20th ACM SIGKDD international conference on Knowledge discovery and data
  mining}.\hskip 1em plus 0.5em minus 0.4em\relax ACM, 2014, pp. 323--332.

\bibitem{lecun1998gradient}
Y.~LeCun, L.~Bottou, Y.~Bengio, and P.~Haffner, ``Gradient-based learning
  applied to document recognition,'' \emph{Proceedings of the IEEE}, vol.~86,
  no.~11, pp. 2278--2324, 1998.

\bibitem{phillips2005overview}
P.~J. Phillips, P.~J. Flynn, T.~Scruggs, K.~W. Bowyer, J.~Chang, K.~Hoffman,
  J.~Marques, J.~Min, and W.~Worek, ``Overview of the face recognition grand
  challenge,'' in \emph{Computer vision and pattern recognition, 2005. CVPR
  2005. IEEE computer society conference on}, vol.~1.\hskip 1em plus 0.5em
  minus 0.4em\relax IEEE, 2005, pp. 947--954.

\bibitem{griffin2007caltech}
G.~Griffin, A.~Holub, and P.~Perona, ``Caltech-256 object category dataset,''
  2007.

\bibitem{wang2014unsupervised}
D.~Wang and X.~Tan, ``Unsupervised feature learning with c-svddnet,''
  \emph{Eprint Arxiv}, 2014.

\end{thebibliography}
\end{document}